\documentclass{article}

\usepackage{arxiv}

\usepackage{soul}
\usepackage[utf8]{inputenc}
\usepackage[T1]{fontenc}
\usepackage{hyperref}
\usepackage{url}
\usepackage{booktabs}
\usepackage{amsfonts}
\usepackage{nicefrac}
\usepackage{microtype}
\usepackage{lipsum}
\usepackage{graphicx}
\usepackage{array}
\usepackage{doi}
\usepackage{multirow}
\usepackage{multicol}
\usepackage{makecell}
\usepackage{xcolor}
\usepackage{xspace}
\usepackage{subcaption}
\usepackage[normalem]{ulem}
\usepackage{amsmath,amsfonts,bm}
\usepackage{amssymb,amsthm}
\usepackage{enumitem}
\usepackage{caption}
\usepackage{adjustbox}
\usepackage{afterpage}
\usepackage[most]{tcolorbox}
\usepackage{listings}
\usepackage{wrapfig} 
\usepackage[table]{xcolor}
\usepackage{pifont}
\usepackage{wrapfig}

\newtcblisting{mybox}[1]{
    colback=gray!5,
    colframe=black!75,
    fonttitle=\bfseries\sffamily,
    title=#1,
    enhanced,
    attach boxed title to top left={yshift=-2mm, xshift=2mm},
    boxed title style={sharp corners=south, colback=black!75},
    rounded corners,
    arc=3pt,
    left=10pt, right=10pt, top=10pt, bottom=10pt,
    listing only,
    listing options={
        basicstyle=\ttfamily\small,
        breaklines=true,
        breakindent=0pt,
        breakautoindent=false,
        columns=fixed,
        keepspaces=true,
        extendedchars=false
    }
}

\usepackage[
    backend=biber,
    citestyle=authoryear,
    bibstyle=authortitle,
    mincitenames=1,
    maxcitenames=1,
    maxbibnames=3
]{biblatex}
\definecolor{hy}{RGB}{255, 128, 0}
\providecolor{BrickRed}{RGB}{184, 20, 11}
\providecolor{OliveGreen}{RGB}{85, 107, 47}

\definecolor{lightgreen}{RGB}{212, 233, 208}
\renewcommand{\thefootnote}{\fnsymbol{footnote}}
\setlist[itemize,1]{leftmargin=18pt, noitemsep, topsep=3pt}
\setlist[enumerate,1]{leftmargin=18pt, noitemsep, topsep=3pt}
\newcommand{\ourname}{PerceptionBench\xspace}
\newcommand{\ournum}{3{,}000\xspace}
\newcommand{\ourmodelname}{{Kimi K3}\xspace}

\renewcommand{\shorttitle}{\raisebox{-0.12\height}{\includegraphics[width=0.02\textwidth]{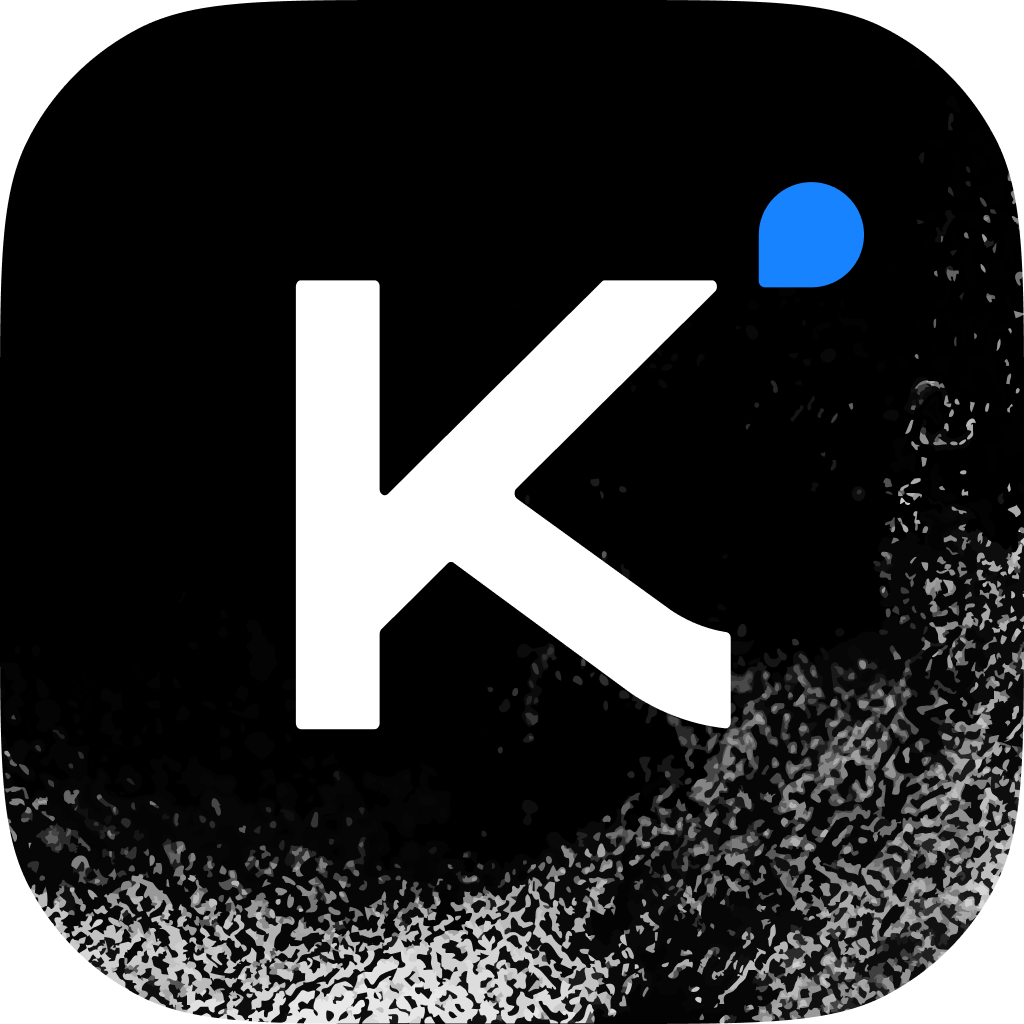}}\hspace{4pt}}

\title{\ourname: Evaluating Atomic Visual Perception \\in Multimodal Large Language Models}

\author{
Moonshot AI
}

\date{}

\begin{document}
\maketitle

\vspace{-10pt}
We introduce \ourname, a benchmark specifically designed to evaluate the atomic visual perception capabilities of Multimodal Large Language Models (MLLMs).
Existing benchmarks often fail to isolate perception: holistic evaluations conflate perceptual errors with failures in reasoning or domain knowledge, while application-driven benchmarks only cover narrow, fragmented domains shaped by heuristic designs.
To address these limitations, \ourname adopts a bottom-up approach: by diagnosing the earliest failure points in the responses of frontier MLLMs across 42 existing benchmarks, we construct an error taxonomy whose perception branch defines ten atomic perceptual capabilities.
Guided by this taxonomy, we construct 3,000 verified questions with short, unambiguous answers, each isolating a single capability, with difficulty stemming from perception rather than reasoning or knowledge. 
Benchmark results across sixteen frontier MLLMs reveal that atomic perception remains largely unsolved---no model reaches 60\% accuracy, perception-related hallucination is the weakest capability on average, and similar overall scores conceal sharply divergent capability profiles.
\ourname thus provides a capability-level standard for measuring and diagnosing the visual perception boundaries of MLLMs.

{\let\thefootnote\relax\footnotetext{Blog: \url{https://www.kimi.com/blog/perception-bench}.}}%
{\let\thefootnote\relax\footnotetext{Code and data: \url{https://github.com/MoonshotAI/PerceptionBench}.}}

\section{Introduction}
\label{sec:intro}

\begin{figure*}[ht]
    \centering
    \includegraphics[width=0.87\textwidth]{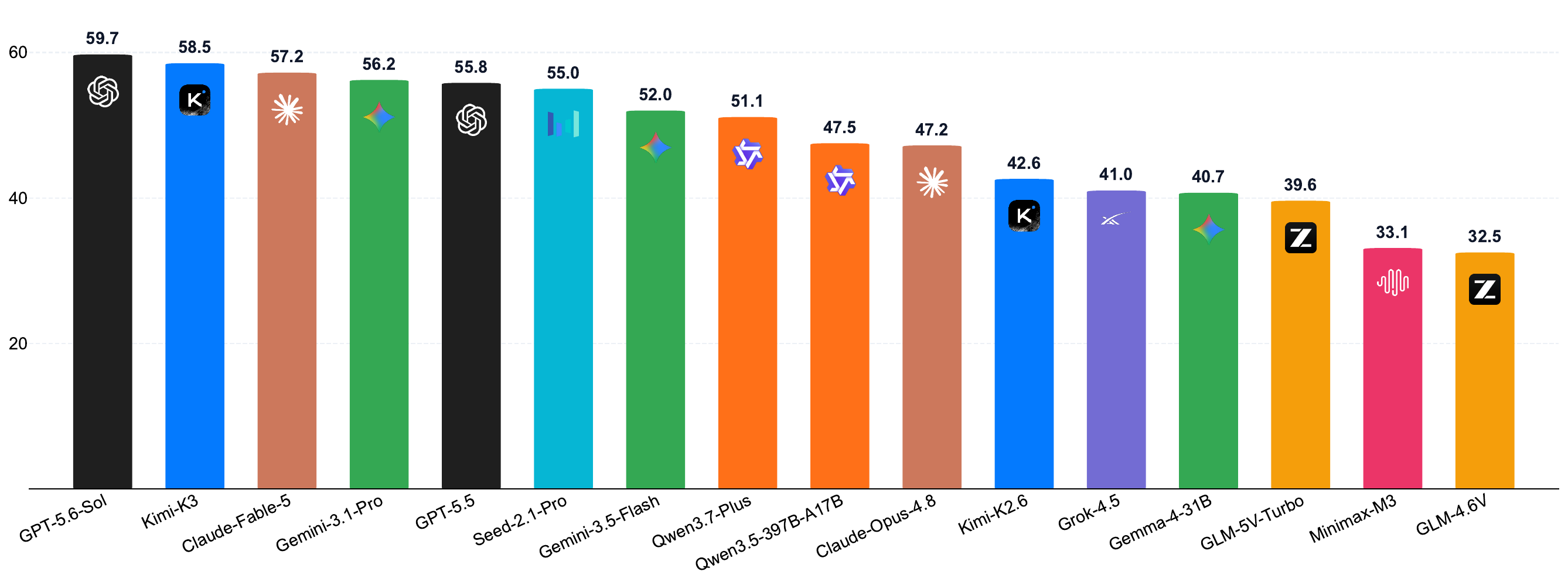}
    \caption{\textbf{Overall model accuracy on \ourname.} While GPT-5.6-Sol (59.7\%) and \ourmodelname (58.5\%) currently lead the field, no evaluated model surpasses the 60\% accuracy threshold, underscoring the significant challenge of atomic visual perception.}
    \label{fig:rank}
\end{figure*}

\begin{figure*}[t]
    \centering
    \includegraphics[width=\textwidth]{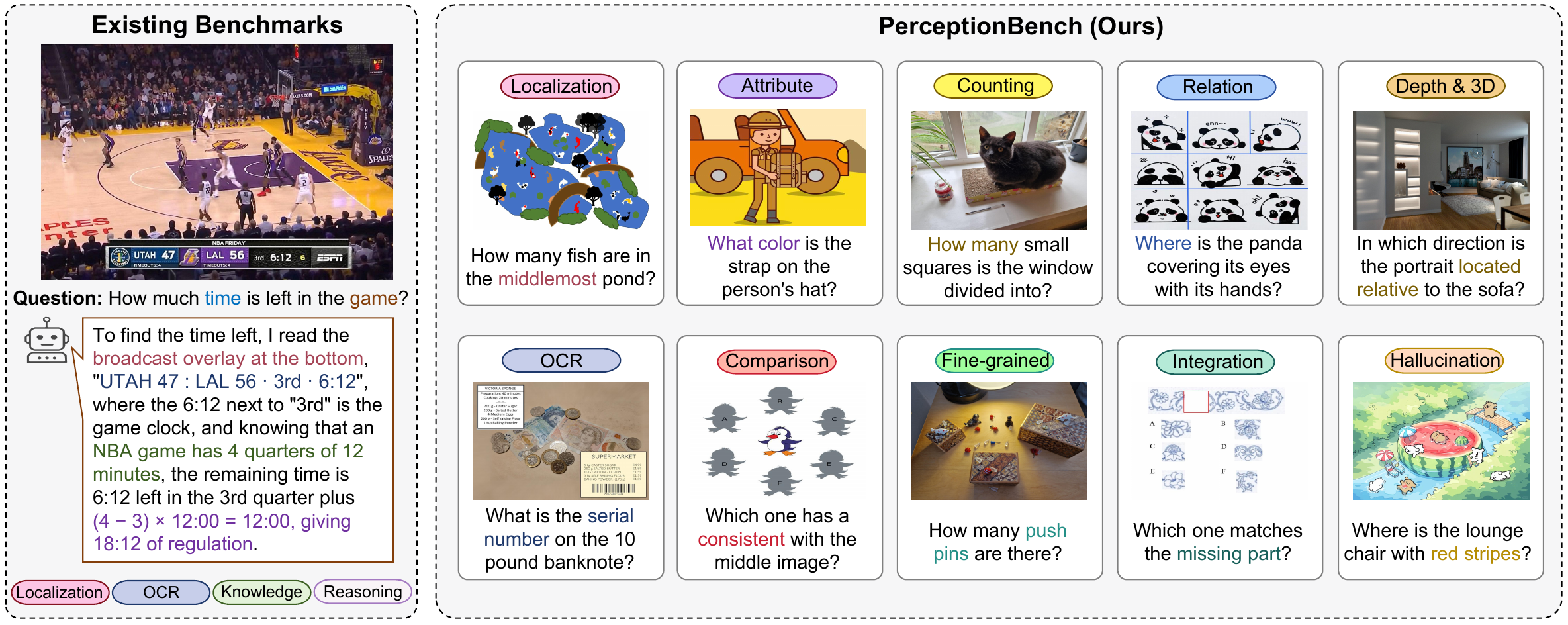}
    \caption{\textbf{\ourname vs.\ existing benchmarks.} \ourname decomposes perception into fine-grained atomic abilities, enabling more detailed and comprehensive evaluation compared to existing benchmarks.}
    \label{fig:teaser}
\end{figure*}

Accurate visual perception serves as a foundational prerequisite for Multimodal Large Language Models (MLLMs) to perform downstream reasoning and environmental interaction~\parencite{fu2024blink, rahmanzadehgervi2025vision, huang2025human}.
However, visual perception is not a single, monolithic ability; it is composed of distinct capabilities---attribute recognition, counting, localization, and text reading---which we term atomic capabilities.
Since downstream reasoning cannot reliably recover visual content that was missed, misread, or hallucinated, uncorrected perceptual errors can invalidate subsequent reasoning~\parencite{schulze2025visual,gao2025pixels}.
Consequently, mastering atomic perception represents a key leverage point for advancing MLLMs.

Despite this foundational role, current evaluations rarely both isolate visual perception from the capabilities built on top of it and systematically cover the range of perceptual capabilities that models lack. Benchmarks such as MMMU~\parencite{yue2024mmmu} and MathVista~\parencite{lu2024mathvista} grade final answers on multi-step problems, where performance depends not only on perception but also on reasoning and external knowledge. Neither a wrong nor a correct answer can therefore be cleanly attributed to perception alone (Figure~\ref{fig:teaser}, left). Meanwhile, benchmarks targeting a single perceptual application, such as OCRBench~\parencite{liu2024ocrbench} and ScreenSpot~\parencite{li2025screenspot}, provide more direct measures of perception but are built top-down: their categories are predefined and populated with questions from a specific application domain. Each benchmark therefore measures only one slice of perception, with its coverage largely shaped by designer priors rather than by the perceptual capabilities that models lack.

We introduce \ourname, a benchmark of \ournum verified questions organized around ten atomic perceptual capabilities (Section~\ref{sec:bench}).
Rather than defining these capabilities from designer priors, we derive them from the failures of frontier MLLMs on 42 existing benchmarks: tracing each failure to its earliest erroneous step and clustering the resulting error labels yields a taxonomy whose perception branch defines the ten capabilities (Section~\ref{subsubsec:discovery}).
Some questions are decomposed from attributed failures into atomic sub-questions, inheriting their capability labels from the attribution stage; others are authored anew against the taxonomy.
Every question is verified so that its difficulty stems from perception rather than reasoning or external knowledge (Figure~\ref{fig:teaser}, right).
Because every capability is grounded in an empirically observed weakness, \ourname directly measures those that models lack, providing a fine-grained diagnosis that task-oriented benchmarks are not designed to offer.

Evaluations of sixteen frontier MLLMs on \ourname show that these models remain far from mastering atomic perception (Figure~\ref{fig:rank}): no model reaches 60\% overall accuracy, and performance is lowest, on average, on perception-related hallucination.
Models with nearly identical overall scores exhibit sharply divergent capability profiles, so aggregate accuracy conceals which perceptual capabilities a model has actually acquired.
A substantial fraction of correct answers are not consistently reproduced across repeated runs (Section~\ref{sec:reliability}), indicating that per-sample perception remains unstable even when aggregate scores are stable.

Our contributions are threefold.
(1) We conduct a failure-attribution study of frontier MLLMs on 42 open-source benchmarks, inducing an error taxonomy whose perception branch defines ten atomic perceptual capabilities, and show that existing benchmarks probe narrow, weakly overlapping slices of the failure space.
(2) We construct \ourname, \ournum verified questions organized around these capabilities and difficulty-balanced so that performance reflects perception rather than reasoning or knowledge.
(3) We evaluate sixteen frontier MLLMs and provide a fine-grained analysis of their perceptual capabilities, showing that atomic perception remains a key bottleneck and that aggregate scores conceal substantial differences in capability profiles.

\section{\ourname}
\label{sec:bench}

\begin{figure*}[t]
    \centering
    \includegraphics[width=\textwidth]{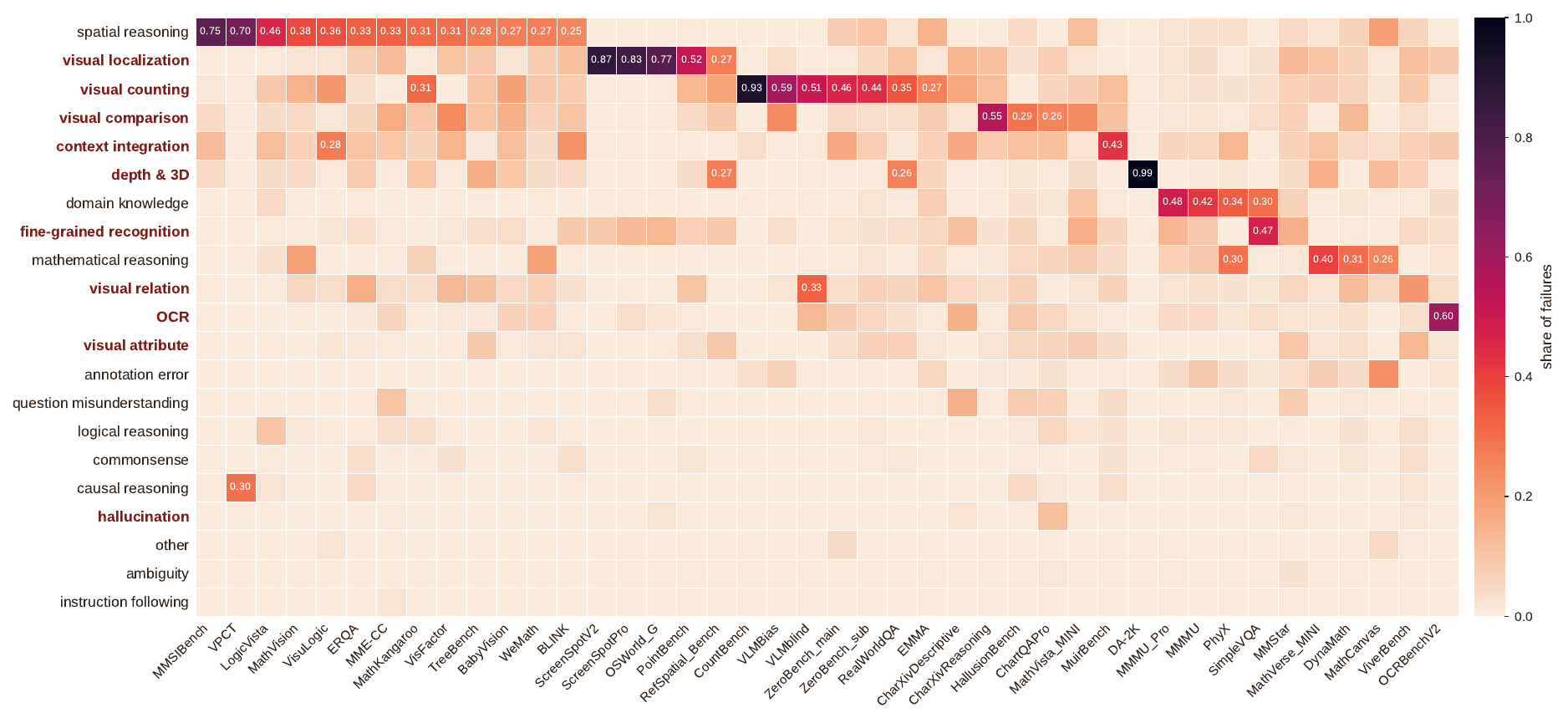}
    \caption{\textbf{Failure-type coverage of existing benchmarks.} Distribution of attributed failures across error types for each of the 42 aggregated open-source benchmarks (rows: error types, sorted by global frequency; columns: benchmarks, ordered by their dominant error type). Each benchmark's failures concentrate on one or a few error types, so existing suites probe narrow, complementary slices of the failure space, while the ten perception-branch types (bold labels) recur across nearly all benchmarks. The never-observed over-refusal type is omitted.}
    \label{fig:coverage}
\end{figure*}

\ourname is designed to identify and evaluate the atomic perceptual capabilities required by MLLMs.
Because each downstream task exercises a combination of these capabilities, each existing benchmark observes only a sparse, application-biased slice of the capability space.
Our failure-attribution study confirms this empirically: failures in each of the 42 aggregated benchmarks concentrate on one or a few error types (Figure~\ref{fig:coverage}), and their error-type distributions overlap only weakly across benchmarks (Figure~\ref{fig:jaccard}).
As a result, within this framework, no individual benchmark or small group of benchmarks covers the space of perceptual failures.
\ourname therefore follows a three-stage bottom-up pipeline that grounds every evaluation category in empirically observed model failures rather than designer priors.
First, we attribute the failures of frontier MLLMs on 42 existing benchmarks to the earliest erroneous step of their reasoning trajectories and cluster the resulting error labels into a unified taxonomy, whose perception branch defines ten atomic perceptual capabilities (Section~\ref{subsubsec:discovery}).
Second, guided by this taxonomy, we select informative failures, decompose them into atomic sub-questions, and author additional questions, balancing the pool across categories and difficulty (Section~\ref{subsubsec:selection}).
Third, every sample undergoes multi-stage verification combining automated capability-alignment and visual-grounding checks with independent human validation (Section~\ref{subsec:bench_5}).
The released benchmark comprises \ournum verified questions, subsampled from the constructed portion---decomposed or newly authored---of an in-house pool of more than 17{,}000 samples (Section~\ref{sec:stats}).

\begin{figure*}[t]
    \centering
    \includegraphics[width=\textwidth]{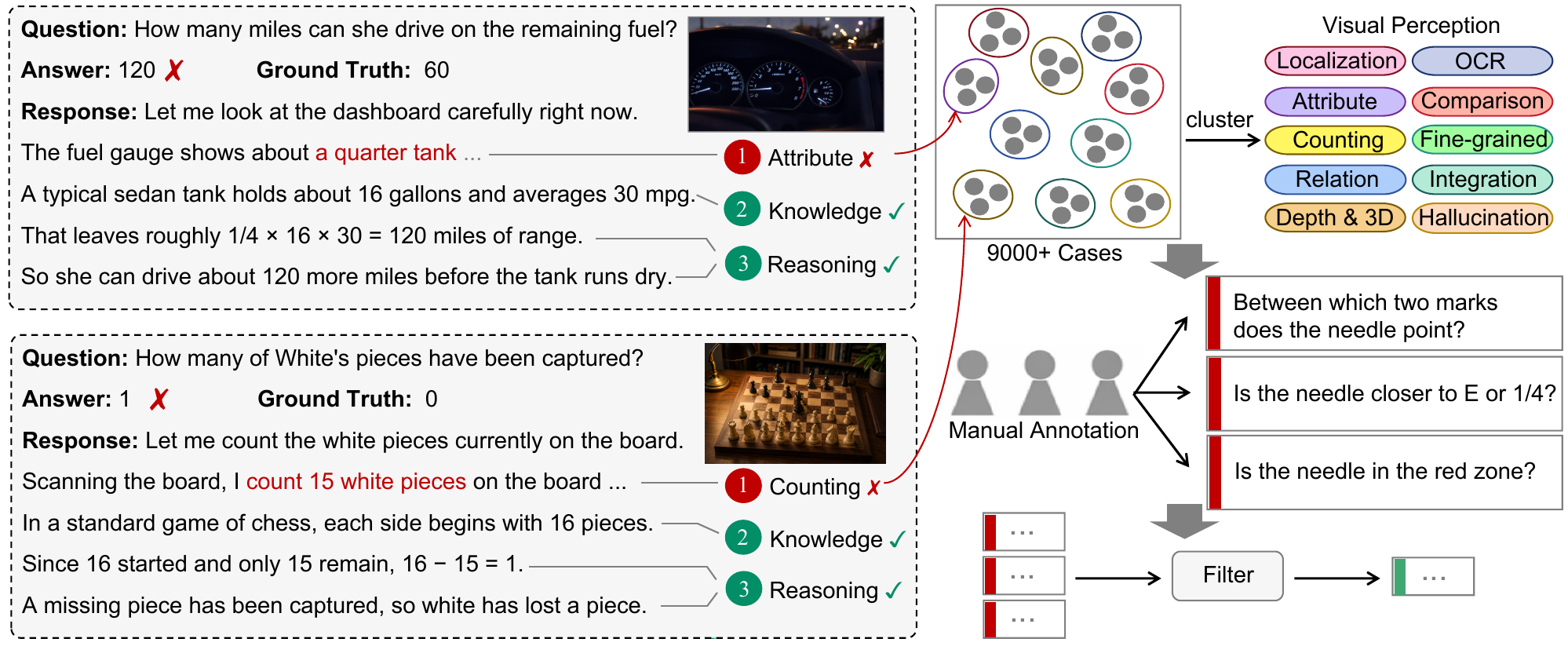}
\caption{\textbf{Construction pipeline of \ourname.} 
(1)~Each model failure on the 42 source benchmarks is attributed to the earliest erroneous step of its reasoning trajectory (\textbf{left});
(2)~cluster the attributed error labels into a unified taxonomy whose perception branch defines the ten atomic perceptual capabilities (\textbf{top right});
(3)~manual annotation produces new perception questions targeting the ten atomic capabilities (\textbf{middle right});
(4)~a final screening filter then removes unsuitable items, yielding the suitable questions of \ourname (\textbf{bottom right}).
}
    \label{fig:pipeline}
\end{figure*}

\subsection{Benchmark Construction}
\label{subsec:construction}
The construction of \ourname proceeds in two stages, followed by the multi-stage verification in Section~\ref{subsec:bench_5}. We first discover a taxonomy of atomic perceptual capabilities from the failures of current MLLMs (Section~\ref{subsubsec:discovery}), and then select and construct evaluation samples that faithfully and comprehensively cover this taxonomy (Section~\ref{subsubsec:selection}).

\subsubsection{Failure-driven Capability Discovery}
\label{subsec:bench_2}
\label{subsubsec:discovery}
Instead of manually defining perceptual capabilities from prior knowledge, we recover this structure from the empirical failures of current MLLMs.
Failures reveal the perceptual phenomena that existing models have not yet reliably acquired, making them substantially more informative than correctly solved examples for understanding perceptual limitations.
This construction is therefore grounded in observed failures rather than solely in designer priors; it remains conditioned on the current generation of models and the coverage of the source benchmarks, and can be re-induced as models evolve.

To obtain these informative failures, we first collect 42 benchmarks spanning diverse visual domains and application scenarios, including document understanding, optical character recognition, chart understanding, GUI interaction, visual grounding, scientific diagrams, remote sensing, and natural images.
Although these benchmarks are designed for different downstream tasks, together they expose a broad range of perceptual phenomena encountered by MLLMs.
We run a set of frontier MLLMs on these benchmarks and collect the samples they answer incorrectly, forming a pool of candidate failures.
However, an incorrect prediction only indicates that the end-to-end inference process has failed, without revealing where the failure occurs: it may originate from visual perception, but also from reasoning, external knowledge, instruction following, or annotation ambiguity.

\label{subsec:bench_3}
To identify failures that genuinely reflect perceptual limitations, we introduce a perceptual failure attribution procedure.
For each candidate failure, we collect the complete prediction context, including the question, image, reference answer, and the model's reasoning trajectory.
A stronger frontier model then analyzes the entire inference process, produces a free-form error analysis that pinpoints the earliest stage responsible for the incorrect answer, and assigns the failure an open-vocabulary error label.
We then cluster and merge semantically equivalent labels with an LLM, consolidating them into a unified taxonomy of recurring error types.
The resulting taxonomy consists of five major error classes, namely perception, reasoning, knowledge, premise, and other errors, which are further divided into 22 fine-grained error types (the complete taxonomy is provided in Appendix~\ref{sec:taxonomy}).
Attribution follows a first-error-priority rule: whenever the error trajectory contains a misread visual fact, the failure is attributed to the earliest erroneous step along the perceptual chain, regardless of whether downstream reasoning also fails; only when all visual facts are correctly extracted is a failure attributed to the reasoning or knowledge classes.

Since \ourname targets visual perception, we retain only the perception error class, whose ten error types we adopt as the ten atomic perceptual capabilities evaluated by \ourname: visual localization, visual attribute recognition, visual counting, visual relation understanding, depth and 3D perception, OCR, visual comparison, fine-grained recognition, context integration, and perception-related hallucination.
The remaining classes characterize non-perceptual limitations and are used only for attribution, not for benchmark construction.

\subsubsection{Benchmark Selection and Construction}
\label{subsubsec:selection}
Guided by the induced taxonomy, we select the most informative failures and construct the evaluation samples of \ourname.

\paragraph{Difficulty-aware selection.}
Directly using the original benchmark distributions is insufficient, as a considerable proportion of existing samples are now consistently solved by frontier MLLMs and contribute little diagnostic value.
We therefore evaluate all candidate samples with an ensemble of four frontier MLLMs.
A sample is discarded as saturated only when it is solved correctly by every ensemble model across all evaluation runs; that is, we remove items judged trivial by the whole ensemble.
The remaining samples are stratified into three difficulty tiers according to the number of models that answer them correctly, and we sub-sample the retained items across source benchmarks and difficulty (pass rate) to counteract the skewed source distribution. 
The four ensemble members are disjoint from the sixteen models evaluated in Section~\ref{sec:settings}.
Difficulty is thus calibrated against current frontier models as a group rather than against any individual model.
We then re-balance the retained perception samples according to the induced error-type distribution, forming the initial in-house pool of \ourname.

\paragraph{Perception decomposition.}
To further smooth the difficulty distribution, we decompose many perception questions from the source benchmarks into finer atomic sub-questions.
For every sample attributed to a perception error type, we provide annotators with its original question, the error analysis generated during attribution, and the assigned perception error type, and ask them to decompose the item into one or more atomic sub-questions, each isolating the specific perceptual capability responsible for the failure.
Every constructed sample is thus grounded in an empirically observed perceptual weakness and inherits its fine-grained capability label from the attribution stage.

\paragraph{Visual data and deduplication.}
To increase visual diversity and improve coverage of underrepresented capabilities, we supplement the benchmark with additional images collected from public web sources and internally curated data, from which annotators author perception questions targeting the same ten capabilities.
To mitigate potential contamination, all images, including both those reused from source benchmarks and those newly collected, are checked against large-scale pre-training corpora (e.g., LAION and Common Crawl).
For each candidate image, we compute the cosine similarity of its Image Similarity Challenge (ISC) descriptor~\parencite{yokoo2021contrastive} against these corpora. Any image whose similarity exceeds a strict threshold of $0.95$, i.e., near-duplicate or leaked images, is discarded.
All source benchmarks are used in accordance with their original licenses (Appendix~\ref{sec:license}).
The decomposition and annotation are performed by more than ten professional annotators with extensive experience in multimodal evaluation.
They ensure that each question admits a unique answer directly grounded in the visual content and that its difficulty arises from perception rather than additional reasoning.

\paragraph{Final screening.}
Finally, mirroring the discovery stage, every newly constructed sample is screened by the same ensemble of frontier MLLMs; samples solved correctly by all of them are removed, and survivors are stratified by difficulty.
This keeps the benchmark challenging for frontier models without collapsing onto either trivial examples or purely adversarial cases, and because the screening is again defined by ensemble consensus, it avoids tailoring the benchmark to any individual model's failure profile.

\paragraph{From the in-house pool to the released benchmark.}
Together, the selected samples and the constructed samples form the complete in-house benchmark of more than 17{,}000 verified samples, which we treat as the full evaluation pool from which the public release is drawn.
The released \ourname consists of \ournum samples subsampled from the \emph{constructed} portion of this pool, with capability-level balancing and difficulty stratification (Section~\ref{sec:stats}).
Section~\ref{sec:represent} further verifies that the released subset preserves the per-capability difficulty structure and relative model performance of the full pool, so evaluating on the released benchmark remains representative at less than one fifth of the full-pool evaluation cost.

\subsection{Data Verification}
\label{subsec:bench_5}
We perform a multi-stage verification process to examine capability alignment, visual grounding, and annotation reliability. Samples that introduce unrelated reasoning requirements, ambiguous visual evidence, or incorrect annotations are removed or revised.

\paragraph{Capability Alignment Verification.}
Each sample is first evaluated by a strong multimodal verifier to determine whether the target perceptual capability is the primary factor required for solving the task.
The verifier analyzes the image, question, answer, and capability annotation, and checks whether the failure of a model on this sample can be attributed to insufficient visual information acquisition rather than reasoning complexity, external knowledge, or instruction misunderstanding.
Samples where the target capability is not sufficiently isolated are excluded or reassigned.

\paragraph{Visual Grounding Verification.}
For each retained sample, we further verify that the answer can be derived from the visual content itself. 
The required information must be either explicitly presented in the image or obtained through integration of multiple visual elements within the image.
Samples requiring external knowledge, unsupported assumptions, or information beyond the visual input are removed to prevent evaluation from being dominated by non-perceptual factors.

\paragraph{Independent Human Validation.}
After automated verification, human annotators independently review the image-question-answer pairs and capability assignments.
Annotators examine whether the visual evidence is sufficient, whether the question matches the intended perceptual challenge, and whether the reference answer is unambiguous.
Samples with inconsistent judgments, unclear visual evidence, or annotation errors are manually corrected or discarded.

\subsection{Benchmark Statistics}
\label{sec:stats}

\begin{figure*}[tb]
\centering
\adjustbox{max width=\textwidth,max totalheight=0.42\textheight}{
\begin{minipage}{0.6\textwidth}
    \centering
    \includegraphics[width=\linewidth]{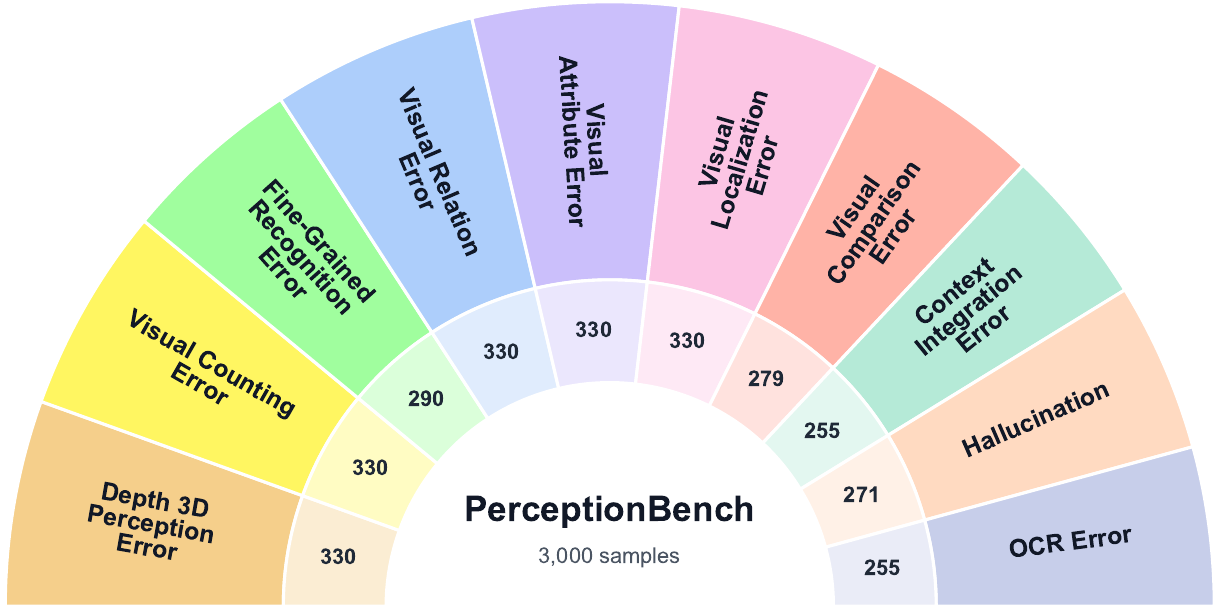}
\end{minipage}
\hfill
\begin{minipage}{0.38\textwidth}
    \centering
    \includegraphics[width=\linewidth]{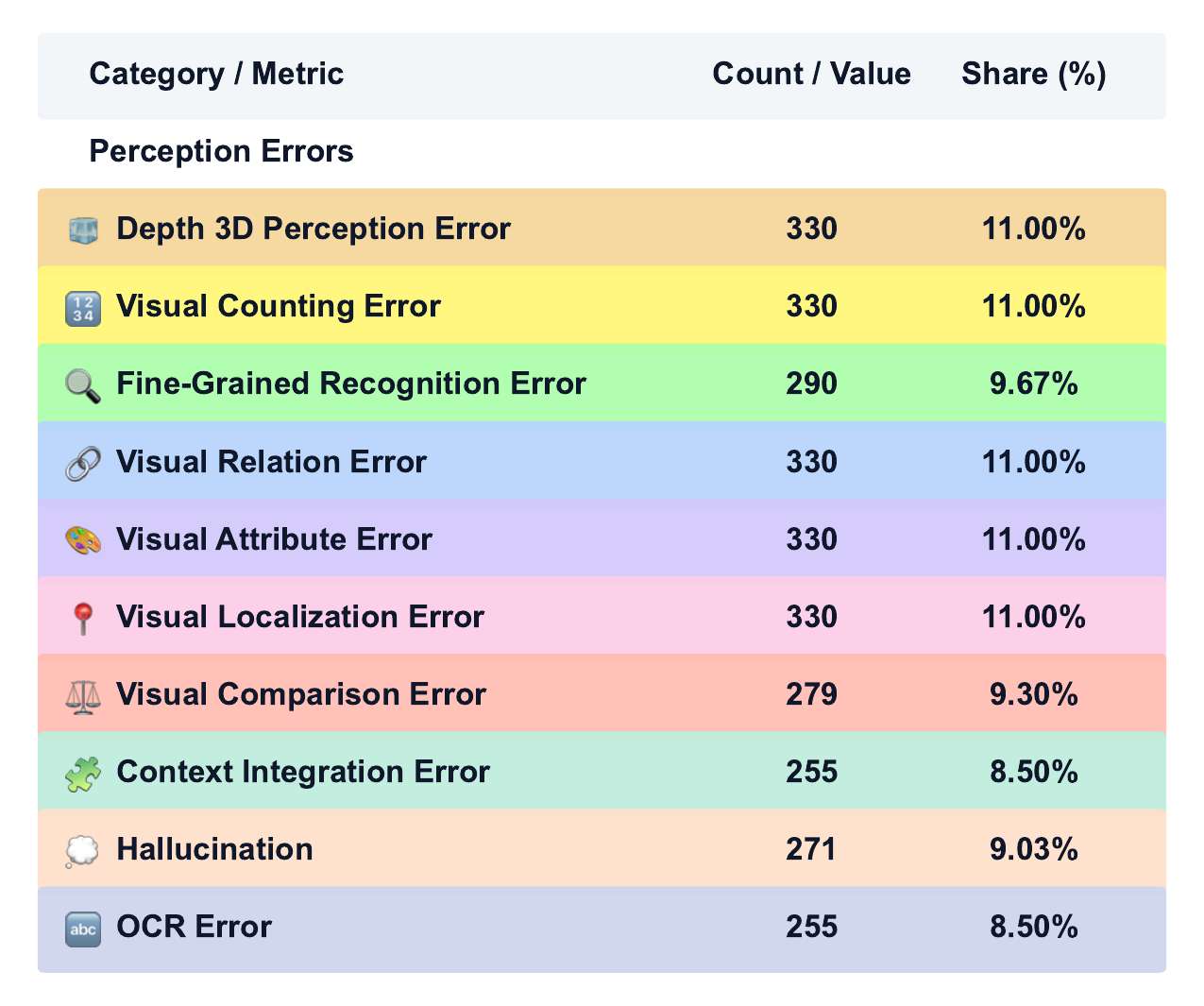}
\end{minipage}
}
\caption{Distribution of \ourname across the ten atomic perceptual capabilities.}
\label{fig:detail}
\end{figure*}

As shown in Figure~\ref{fig:detail}, \ourname~contains \ournum verified samples spanning the ten atomic perceptual capabilities identified from the failure-driven capability discovery process.
In terms of construction source, 1{,}800 of the \ournum questions (60\%) are atomic sub-questions decomposed from attributed failures on the source benchmarks, while the remaining 1{,}200 (40\%) are newly authored on supplemented images (Section~\ref{subsubsec:selection}).
By design, the remaining error classes of the induced taxonomy are used exclusively for failure attribution during construction and are excluded from the released benchmark, keeping the evaluation focused on visual perception itself.
Overall, the benchmark distribution follows the empirical frequency of perceptual failures observed in current MLLMs while applying capability-level balancing to prevent dominant failure modes from overwhelming the evaluation, and samples are further stratified by difficulty. This design enables \ourname to provide both comprehensive coverage of visual perception capabilities and fine-grained analysis of model limitations.

\section{Evaluation Results}

\begin{table*}[t]
\centering
\resizebox{\textwidth}{!}{
\begin{tabular}{l|c|cccccccccc}
\toprule
\textbf{Model} & \textbf{Overall} & \textbf{VRel} & \textbf{Count} & \textbf{Attr} & \textbf{Depth} & \textbf{Loc} & \textbf{Comp} & \textbf{FGR} & \textbf{Context} & \textbf{OCR} & \textbf{Hallu}
\\
\midrule
\rowcolor{gray!10}
\multicolumn{12}{c}{\textbf{Proprietary MLLMs}}
\\
\midrule
GLM-5V-Turbo & 39.6 & 41.2 & 40.0 & 41.2 & 41.2 & 43.9 & 45.2 & 36.6 & 32.9 & 43.5 & 28.0 \\
Grok-4.5 & 41.0 & 47.0 & 35.2 & 39.4 & 41.2 & 39.7 & 43.7 & 36.2 & 39.6 & 43.9 & 44.7 \\
Claude-Opus-4.8 & 47.2 & 51.4 & 44.2 & 49.4 & 40.6 & 58.8 & 48.4 & 40.7 & 44.7 & 54.1 & 38.7 \\
Qwen3.7-Plus & 51.1 & 59.1 & 53.3 & 55.8 & 48.5 & 52.7 & 55.9 & 46.8 & 52.2 & 54.5 & 29.5 \\
Gemini-3.5-Flash & 52.0 & 53.6 & 43.6 & 54.5 & 49.7 & 50.6 & 54.8 & 51.7 & 53.3 & 59.6 & 50.6 \\
Seed-2.1-Pro & 55.0 & 57.6 & 51.2 & 58.2 & 43.6 & 50.0 & 59.5 & 56.6 & 60.4 & 66.7 & 49.8 \\
GPT-5.5 & 55.8 & 61.9 & 55.8 & 60.9 & 48.8 & 65.8 & 65.6 & 47.2 & 58.0 & 56.5 & 34.7 \\
Gemini-3.1-Pro & 56.2 & 58.8 & 56.9 & 61.8 & 50.0 & 52.7 & 61.7 & 54.8 & 61.2 & 64.3 & 40.6 \\
Claude-Fable-5 & 57.2 & 58.5 & 52.9 & 60.9 & 51.5 & 70.4 & 56.1 & 51.6 & 59.8 & 64.3 & 45.0 \\
GPT-5.6-Sol & \textbf{59.7} & 69.7 & 62.4 & 62.1 & 55.5 & 76.7 & 67.0 & 55.9 & 60.0 & 54.9 & 26.9 \\
\midrule
\rowcolor{gray!10}
\multicolumn{12}{c}{\textbf{Open-source MLLMs}}
\\
\midrule
GLM-4.6V & 32.5 & 35.2 & 31.8 & 35.2 & 29.1 & 30.6 & 34.8 & 29.3 & 33.7 & 39.2 & 26.9 \\
Minimax-M3 & 33.1 & 40.0 & 30.3 & 34.6 & 36.7 & 33.3 & 31.2 & 26.6 & 31.0 & 35.7 & 29.9 \\
Gemma-4-31B & 40.7 & 42.7 & 33.9 & 40.3 & 39.1 & 44.9 & 43.7 & 39.0 & 45.9 & 46.7 & 32.1 \\
Kimi K2.6 & 42.6 & 50.9 & 45.2 & 43.6 & 42.4 & 45.2 & 39.1 & 34.5 & 40.4 & 40.8 & 41.0 \\
Qwen3.5-397B-A17B & 47.5 & 55.2 & 49.1 & 53.0 & 44.6 & 46.7 & 49.8 & 44.8 & 50.2 & 52.9 & 26.9 \\
\ourmodelname & \textbf{58.5} & 68.2 & 59.7 & 59.4 & 52.4 & 70.3 & 59.1 & 55.9 & 53.3 & 61.2 & 41.7 \\
\bottomrule
\end{tabular}
}
\caption{
\textbf{Leaderboard evaluation on \ourname.}
The benchmark evaluates MLLMs across ten atomic perceptual capabilities identified from failure-driven capability induction.
VRel, Count, Attr, Depth, Loc, Comp, FGR, Context, OCR, and Hallu denote visual relation, counting, attribute, depth and 3D perception, localization, comparison, fine-grained recognition, context integration, optical character recognition, and perception-related hallucination, respectively.}
\label{tab:leaderboard}
\end{table*}

\subsection{Experiment Settings}
\label{sec:settings}
To ensure a rigorous and fair evaluation across the diverse landscape of MLLMs, we maintain strict consistency in our experimental protocols: all models are evaluated with unified prompts and official inference settings where available.
We evaluate a suite of representative frontier MLLMs spanning both proprietary and open-source families, including ten proprietary models (Claude-Opus-4.8, Claude-Fable-5, GPT-5.6-Sol, GPT-5.5, Gemini-3.5-Flash, Gemini-3.1-Pro, Qwen3.7-Plus, Seed-2.1-Pro, GLM-5V-Turbo, and Grok-4.5) and six open-source models (\ourmodelname, Kimi K2.6, Qwen3.5-397B-A17B, Gemma-4-31B, Minimax-M3, and GLM-4.6V).
For each model, we adopt the highest available reasoning budget. Specifically, GPT-5.5 uses ``xhigh'' mode, GPT-5.6-Sol, Claude-Fable-5, Claude-Opus-4.8 and \ourmodelname use ``max'' mode, the Gemini series and Seed-2.1-Pro use ``high'' effort, and Gemma-4-31B runs with thinking mode enabled. Claude-Fable-5 is evaluated with Claude-Opus-4.8 as a fallback on queries refused by its safety filter.
In total, 1.1\% of its reported answers are produced by Claude-Opus-4.8.
All questions in \ourname are open-ended free-form questions whose reference answers are short and uniquely determined.
We therefore employ GPT-oss-120B as a unified judge model that compares each model response against the reference answer for automatic evaluation (see Appendix~\ref{sec:prompt} for the evaluation prompt).
On a random sample of 300 predictions, the automatic evaluation agrees with human judgments on 299 cases (99.7\%), which is consistent with the design that reference answers are short and uniquely determined.

\begin{figure*}[t]
    \centering
    \includegraphics[width=\textwidth]{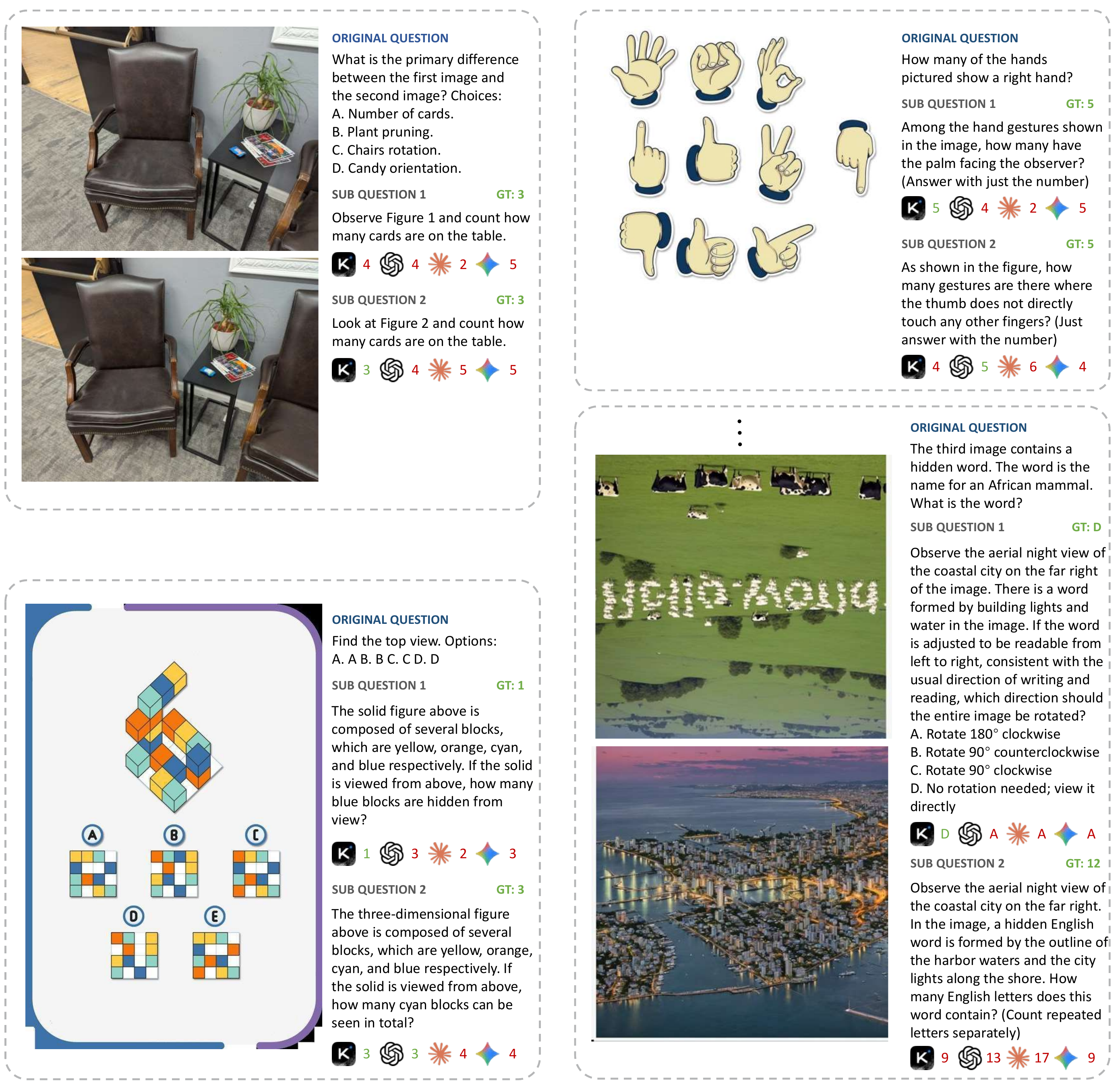}
    \caption{\textbf{Qualitative results on \ourname.} Four source-benchmark items whose original questions require multi-step solutions, decomposed into atomic, perception-only questions (GT: ground truth; colored icons: answers of \ourmodelname, GPT-5.6-Sol, Claude-Fable-5, and Gemini-3.1-Pro). Sources: ERQA~\parencite{gemini2025robotics} (top left), MathVision~\parencite{wang2024mathvision} (top right), MMStar~\parencite{chen2024we} (bottom left), and ZeroBench~\parencite{roberts2025zerobench} (bottom right).}
    \label{fig:qualitative}
\end{figure*}

\subsection{Main Results}
\textbf{Atomic perception remains unsolved.}
As detailed in Table~\ref{tab:leaderboard}, GPT-5.6-Sol leads at 59.7\%, followed closely by \ourmodelname~(58.5\%), the top-performing open-source model, which surpasses all remaining proprietary systems including Gemini-3.1-Pro (56.2\%) and GPT-5.5 (55.8\%).
On questions that require only the faithful acquisition of visual information, this shortfall underscores that atomic perception remains a key bottleneck of current MLLMs.
Overall accuracy spans 32.5\% (GLM-4.6V) to 59.7\%, so the benchmark retains strong discriminative power far from any ceiling effect.

\textbf{Perceptual competence is unevenly developed.}
Across models, perception-related hallucination has the lowest average accuracy (36.7\%), well below visual relation (53.2\%), OCR (52.4\%), and localization (52.0\%).
This weakness is not limited to weaker models: GPT-5.6-Sol, the overall leader, reaches 76.7\% on localization but only 26.9\% on hallucination, one of the lowest hallucination scores in the leaderboard.
Models with nearly identical overall scores also diverge sharply: GPT-5.5 (55.8\%) and Gemini-3.1-Pro (56.2\%) differ by less than one point overall, yet GPT-5.5 leads in localization by 13 points (65.8 vs.\ 52.7), while Gemini-3.1-Pro leads in OCR by 8 points (64.3 vs.\ 56.5) and in fine-grained recognition by 8 points (54.8 vs.\ 47.2).

\textbf{Perceptual capability and reliability decouple.}
The hallucination capability exposes an instructive inversion: the overall leader, GPT-5.6-Sol, records one of the lowest hallucination scores (26.9\%), whereas Gemini-3.5-Flash, which sits mid-pack overall (52.0\%), achieves the best score (50.6\%), closely followed by Seed-2.1-Pro (49.8\%).
We hypothesize that models tuned for aggressive answer commitment tend to assert plausible but non-existent visual content, inflating their scores on answerable questions while failing precisely on the samples designed to probe false-positive perception.
This observation is echoed by the pass@4 versus pass\textsuperscript{4} analysis in Section~\ref{sec:reliability}.

\textbf{The open-source frontier is closing in.}
\ourmodelname~trails the best proprietary model by only 1.2 points while outperforming every other proprietary system.
Within the open-source group, perception scales with model capacity (Qwen3.5-397B-A17B at 47.5\% vs.\ GLM-4.6V at 32.5\%), yet even the strongest models leave every capability below 80\%.

\subsection{Qualitative Analysis}
\label{sec:qualitative}
We further inspect individual inference samples of four frontier models (\ourmodelname, GPT-5.6-Sol, Claude-Fable-5, and Gemini-3.1-Pro) on \ourname.
Figure~\ref{fig:qualitative} shows four representative cases, each decomposed from a single source-benchmark item.
Although the original questions admit multi-step solutions, the models already err on many of the decomposed atomic questions.
These errors are directly attributable to specific perceptual capabilities, such as counting instances or localizing objects, rather than to reasoning or knowledge.
The cases echo the aggregate results: on \ourname, failures are often perceptual before they are cognitive.

\subsection{Benchmark Representativeness}
\label{sec:represent}

\begin{wrapfigure}[23]{R}{0.5\textwidth}
    \centering
    \vspace{-4em}
    \includegraphics[width=\linewidth]{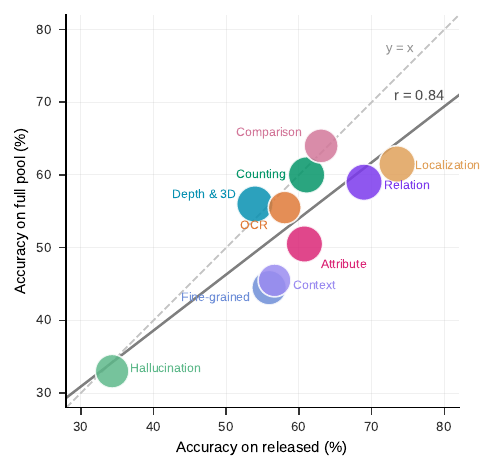}
    \caption{\textbf{Per-capability representativeness of the released \ourname.}
    Each point represents one perceptual capability. The $x$- and $y$-axes show accuracy on the released and full benchmarks, respectively, with point size proportional to the number of samples.}
    \label{fig:represent}
    \vspace{-1em}
\end{wrapfigure}

The released benchmark is subsampled from the constructed samples of the complete in-house benchmark, which contains more than 17{,}000 verified samples.
To evaluate whether the released version faithfully preserves the evaluation characteristics of the full in-house benchmark, we compare per-capability accuracies on the released subset against those on the full benchmark.

As shown in Figure~\ref{fig:represent}, the two are strongly correlated across the ten atomic perceptual capabilities (Pearson $r=0.84$), and the full benchmark is slightly harder than the released subset for most capabilities, as the released version smooths the difficulty distribution.
This suggests that the released subset preserves the per-capability difficulty structure and relative model performance of the full benchmark while substantially reducing evaluation cost.

\subsection{Evaluation Reliability}
\label{sec:reliability}
Since MLLMs may exhibit stochastic behaviors during inference, we evaluate several frontier models using four independent runs.
For each model, we report the accuracy of each run together with the mean and standard deviation.
In addition, we report pass@4 (the proportion of samples solved in at least one of the four runs) and pass\textsuperscript{4} (the proportion of samples solved in all four runs), which characterize the capability upper bound and the behavioral consistency of each model, respectively.

The observed standard deviations remain consistently small (Table~\ref{tab:reliability}): for example, the four runs of \ourmodelname~span only 0.6 points (58.2--58.8, std 0.32), indicating that benchmark scores are stable and insensitive to sampling randomness, and that \ourname~provides reliable evaluation despite the diversity of visual tasks.
Meanwhile, the gap between pass@4 and pass\textsuperscript{4} reveals that a considerable fraction of samples are solved only intermittently across runs: although the aggregate score is stable, per-sample perceptual behavior is not.
This suggests that current models have not robustly acquired the underlying perceptual capabilities, and part of their measured accuracy stems from unstable predictions rather than reliable visual perception.

\begin{table}[!tb]
\centering
\setlength{\tabcolsep}{5pt}
\begin{tabular}{lcccccccc}
\toprule
\textbf{Model} & \textbf{Run1} & \textbf{Run2} & \textbf{Run3} & \textbf{Run4} & \textbf{Mean} & \textbf{Std} & \textbf{pass@4} & \textbf{pass\textsuperscript{4}} \\
\midrule
GPT-5.6-Sol  & 59.9 & 59.5 & 59.8 & 59.8 & 59.7 & 0.17 & 72.9 & 46.9 \\
Gemini-3.5-Flash  & 51.4 & 51.5 & 52.5 & 52.6 & 52.0 & 0.64 & 70.2 & 32.6 \\
\ourmodelname  & 58.2 & 58.2 & 58.8 & 58.7 & 58.5 & 0.32 & 73.9 & 42.7 \\
\bottomrule
\end{tabular}
\vspace{0.2cm}
\caption{\textbf{Evaluation reliability on representative frontier MLLMs.} Each model is evaluated with four independent runs.
We report the accuracy of each run together with the mean and standard deviation, as well as pass@4 (solved in at least one run) and pass\textsuperscript{4} (solved in all runs). Lower standard deviation indicates higher evaluation stability, while the gap between pass@4 and pass\textsuperscript{4} reflects the behavioral inconsistency of model perception.}
\label{tab:reliability}
\end{table}

\section{Related Work}
\label{sec:related}

\paragraph{General-purpose MLLM benchmarks.}
Evaluation of MLLMs has largely been organized around tasks rather than perceptual capabilities.
General-purpose benchmarks grade end-to-end answers on broad task collections.
Early VQA suites~\parencite{antol2015vqa, hudson2019gqa, zhu2016visual7w} test scene understanding through question answering, while recent holistic benchmarks~\parencite{yue2024mmmu, liu2024mmbench, chen2024we, yue2025mmmupro} extend this paradigm to college-level reasoning and knowledge-intensive problems.
Because performance on these benchmarks jointly depends on perception, reasoning, and external knowledge, their aggregate scores do not reveal which of these components accounts for a model's success or failure.

\paragraph{Application-oriented visual benchmarks.}
A second line of work moves closer to perception by targeting individual visual applications, including text recognition~\parencite{liu2024ocrbench, fu2026ocrbench, ma2024mmlongbench}, structured document and chart understanding~\parencite{mathew2021docvqa, masry2022chartqa, mathew2022infographicvqa, wang2024charxiv}, GUI interaction~\parencite{li2025screenspot, xie2024osworld}, spatial interaction~\parencite{gao2026spatialworld}, and visual grounding~\parencite{yu2016modeling, krishna2017visual, cheng2025pointarena}.
These suites probe perception more directly, yet each samples only the perceptual skills exercised by its target application, and many items still require domain-specific reasoning or specialized output formats.
As a result, task-oriented benchmarks measure perception only indirectly and through narrow, imbalanced slices of the perceptual space.
We quantify this fragmentation using the per-benchmark error-type distributions in Figure~\ref{fig:coverage}.

\paragraph{Perception-centric benchmarks.}
A more recent line of work evaluates visual perception directly using short questions designed to minimize reasoning and knowledge demands.
BLINK~\parencite{fu2024blink}, TET~\parencite{gao2025pixels}, and VisFactor~\parencite{huang2025human} assemble core visual cognition tasks that are relatively easy for humans but challenging for MLLMs.
Similarly,~\textcite{rahmanzadehgervi2025vision} evaluate MLLMs using deliberately simple visual questions.
Other benchmarks isolate specific perceptual phenomena, including visual illusion and hallucination~\parencite{guan2024hallusionbench}, multi-image spatial reasoning~\parencite{yang2025mmsibench}, infant-level visual understanding~\parencite{chen2026babyvision}, and extremely difficult visual queries~\parencite{roberts2025zerobench}.
Together, these studies show that frontier MLLMs continue to struggle with basic visual perception.
However, their categories are defined top-down from designer priors, and each benchmark focuses on a particular slice of perception.
Our failure-attribution study further shows that the failures exposed by these benchmarks overlap only weakly (Appendix~\ref{sec:jaccard}), suggesting that each captures a fragment of the perceptual space rather than its broader structure.

\ourname differs from these lines of work primarily in how its evaluation categories are obtained.
Rather than adopting task boundaries or designer-defined categories, it derives ten atomic perceptual capabilities bottom-up from the attributed failures of frontier MLLMs on 42 existing benchmarks.
The question pool is then balanced across these capabilities and difficulty tiers.
In this way, \ourname retains the directness of perception-centric evaluation while grounding its capability coverage in empirically observed model weaknesses.
This design provides a level of diagnostic granularity that task-oriented benchmarks are not designed to offer.

\section{Conclusion}
\label{sec:con}
In this work, we introduce \ourname, a benchmark for evaluating atomic visual perception in MLLMs.
Its ten capability categories are derived bottom-up from attributed failures on 42 existing benchmarks, and each of its \ournum verified questions isolates a single capability.
Evaluations of sixteen frontier MLLMs suggest that atomic perception remains largely unsolved.
No model reaches 60\% overall accuracy, and even the strongest models leave every capability below 80\%.
Perception-related hallucination is the weakest capability on average, including for the top-ranked systems.
Models with nearly identical overall scores exhibit sharply divergent capability profiles, so aggregate accuracy conceals which capabilities a model has actually acquired.
Aggregate scores are also stable across repeated runs while per-sample behavior is not, suggesting that part of the measured accuracy stems from unstable predictions rather than reliable perception.
The open-source frontier is closing in: \ourmodelname trails the overall leader by only 1.2 points.

\section*{Limitations}
Our taxonomy is induced from the failures of the current model generation, so it reflects today's weaknesses rather than a fixed structure of perception.
As models improve, the taxonomy should be re-induced and the ensemble-based difficulty calibration recalibrated.
Failure attribution also relies on a stronger analyzer model, and residual attribution errors may propagate into capability labels.
A related caveat is that each question's capability label is inherited from first-error attribution on the current model generation.
When the first erroneous step of frontier models shifts, the same item may be attributed to a different category.
Finally, \ourname isolates perception by design, so capability scores do not directly predict end-to-end task performance; the benchmark is best used alongside task-oriented suites.
We hope that this capability-level diagnosis guides more targeted improvements of visual perception, and that the failure-driven pipeline allows the benchmark to evolve together with the models it measures. 
\newpage
\printbibliography[title={References}]
\newpage
\appendix
\newpage
\section{Contributions}
{\small
Zichao Lin$^{*}$ \quad
Yifeng Xie$^{*}$ \quad
Bowen Qu$^{*}$ \quad
Haiming Wang \quad
Jia Li \quad
Haoning Wu \quad
Yuhao Dong \quad
Zuhao Yang \quad
Jinguo Zhu \quad \\
Haoyu Lu \quad
Zijia Zhao \quad
Tongtian Yue \quad
Zhangyang Qi \quad
Junwei Yang \quad
Mengfan Dong \quad
Peizhou Cao \quad
Chenzhuang Du \quad \\
Zaida Zhou \quad
Haotian Yao \quad
Hao Yang \quad 
Hongcheng Gao \quad
Lin Sui \quad
Weihong Li \quad
Xinxing Zu \quad
Jia Chen \quad
Yao Wang \quad \\
Xiaoxue Wu \quad
Yalin Wang \quad
Y. Charles \quad
Yiping Bao \quad
Yangyang Liu \quad
Zhiqi Huang$^{\dagger}$ \quad
Xinyu Zhou \quad

\vspace{6pt}
\noindent$^{*}$Core contributors.

\noindent$^{\dagger}$Project lead.
}

\section{Full Error Attribution Taxonomy}
\label{sec:taxonomy}

This appendix presents the complete error attribution taxonomy derived from the failure analysis described in Section~\ref{subsec:bench_3}. The taxonomy comprises five high-level error classes and 22 fine-grained error types, providing a comprehensive categorization of failure sources in multimodal visual understanding. Among them, the perception-related errors constitute the capability taxonomy adopted by \ourname, while the remaining error classes are introduced solely to facilitate error attribution during taxonomy induction.

Table~\ref{tab:taxonomy_perception} summarizes the ten perception error types that define the atomic perceptual capabilities evaluated by \ourname. Table~\ref{tab:taxonomy_other} presents the remaining four error classes, namely reasoning, knowledge, premise, and other errors. These error types are used to distinguish non-perceptual failures during taxonomy induction, but are not included in the released benchmark.

\paragraph{Attribution rule (first-error priority).}
Whenever the error trajectory contains a misread visual fact, the failure is attributed to the \emph{first} erroneous step along the perceptual chain, and mapped to the corresponding perception error type, regardless of whether downstream reasoning is also incorrect. Only when all visual facts in the trajectory are verifiably extracted correctly can the failure be attributed to the reasoning or knowledge classes.

\begin{table}[!htbp]
\centering
\small
\begin{tabular}{p{4.2cm}p{9.6cm}}
\toprule
\textbf{Perception Error Types} & \textbf{Definition and Disambiguation Criteria} \\
\midrule
Visual localization & Fails to locate or anchor the correct region or object in the image. (Misidentifying an object as something else belongs to \emph{fine-grained recognition}.) \\
Visual attribute & Misjudges low-level attributes of a single object, such as color, shape, size, or texture. Attribute comparison across objects belongs to \emph{visual comparison}. \\
Visual counting & Miscounts the number of objects or instances. \\
Visual relation & Misreads directly visible 2D spatial relations (left/right, above/below, inside/outside, etc.). \\
Depth \& 3D perception & Errors in perceiving stereoscopic information such as depth ordering, occlusion, viewpoint or orientation, and 3D shape structure. \\
OCR & Fails to correctly read text, digits, or symbols in the image. \\
Visual comparison & Draws wrong conclusions when comparing two or more visual elements (larger/smaller, same/different, etc.). \\
Fine-grained recognition & Misrecognizes the category of an object or entity, or fails to distinguish visually similar categories, subtypes, and fine-grained differences. \\
Context integration & Each individual region or sub-figure is perceived correctly, but associating and aggregating information across regions, sub-figures, or images fails. \\
Hallucination & Asserts objects, text, or values that do not exist in the image. (If a corresponding real element exists but is misread, the failure belongs to the corresponding specific perception error type.) \\
\bottomrule
\end{tabular}
\vspace{0.2cm}
\caption{Perception error types in the induced error attribution taxonomy. These ten error types define the atomic perceptual capabilities evaluated by \ourname.}
\label{tab:taxonomy_perception}
\end{table}

\begin{table}[!htbp]
\centering
\small
\begin{tabular}{p{4.2cm}p{9.6cm}}
\toprule
\textbf{Error Type} & \textbf{Definition and Disambiguation Criteria} \\
\midrule
\multicolumn{2}{l}{\textbf{Reasoning Errors}} \\
\midrule
Spatial reasoning & Spatial information itself is perceived correctly, but mental transformation or simulation of the spatial structure fails (rotation, folding, imagined viewpoint change, path planning, etc.). Directly misreading a visible relation belongs to \emph{visual relation}; misperceiving depth or viewpoint belongs to \emph{depth \& 3D perception}. \\
Logical reasoning & All visual facts are extracted correctly (verifiable in the trajectory), but pure logical inference or deduction fails. \\
Mathematical reasoning & All values involved in the computation are read correctly, but pure arithmetic, geometric, or quantitative computation fails. Misreading a value itself belongs to \emph{OCR}. \\
Causal reasoning & Facts are extracted correctly, but causal or temporal inference is wrong. \\
\midrule
\multicolumn{2}{l}{\textbf{Knowledge Errors}} \\
\midrule
Domain knowledge & Lacks specialized domain knowledge (science, medicine, law, art, etc.). \\
Commonsense & Lacks everyday commonsense about objects, behaviors, or situations. \\
\midrule
\multicolumn{2}{l}{\textbf{Premise Errors}} \\
\midrule
Question misunderstanding & Misunderstands the question itself before any visual step: wrong referent, missed constraints, or misinterpreted task requirements. \\
Instruction following & Understands the task correctly, but the output format, length, or structure violates the requirements. \\
Over-refusal & Refuses to answer or excessively hedges although a definite answer exists. \\
\midrule
\multicolumn{2}{l}{\textbf{Others}} \\
\midrule
Ambiguity & The question admits multiple reasonable answers; the model answer is reasonable but inconsistent with the ground truth. \\
Annotation error & The ground truth itself is wrong; the model answer may actually be correct. \\
Other & Does not fall into any of the categories above. \\
\bottomrule
\end{tabular}
\vspace{0.2cm}
\caption{Non-perceptual error types in the induced taxonomy. These error types are used for failure attribution during taxonomy induction and are excluded from \ourname by design.}
\label{tab:taxonomy_other}
\end{table}

\section{Source Benchmark List and License}
\label{sec:sources}

Table~\ref{tab:source_benchmarks} lists the 42 open-source benchmarks (including benchmark splits) aggregated for the failure-driven capability discovery pipeline described in Section~\ref{subsec:bench_2}. These benchmarks span diverse visual tasks and application domains, and after ensemble-based difficulty filtering they yield approximately 9,000 informative failure cases, which serve as the basis for inducing the error attribution taxonomy and identifying the atomic perceptual capabilities evaluated in \ourname.

\begin{table}[!htbp]
\centering
\small
\resizebox{\linewidth}{!}{
\begin{tabular}{lll}
\toprule
\textbf{Benchmark} & \textbf{Benchmark} & \textbf{Benchmark} \\
\midrule
MathVision~\parencite{wang2024mathvision} & ChartQAPro~\parencite{masry2025chartqapro} & ERQA~\parencite{gemini2025robotics} \\
PhyX~\parencite{shen2025phyx} & HallusionBench~\parencite{guan2024hallusionbench} & MuirBench~\parencite{wang2024muirbench} \\
VLMBias~\parencite{vo2025vlmbias} & ScreenSpot-Pro~\parencite{li2025screenspot} & We-Math~\parencite{qiao2025wemath} \\
OCRBench-v2~\parencite{fu2026ocrbench} & EMMA~\parencite{hao2025emma} & LogicVista~\parencite{xiao2024logicvista} \\
VisFactor~\parencite{huang2025human} & MathCanvas~\parencite{shi2026mathcanvas} & VLMs-are-Blind~\parencite{rahmanzadehgervi2025vision} \\
PointBench~\parencite{cheng2025pointarena} & BabyVision~\parencite{chen2026babyvision} & CharXiv-Descriptive~\parencite{wang2024charxiv} \\
ViVerBench~\parencite{zhang2025omniverifier} & BLINK~\parencite{fu2024blink} & ZeroBench-main~\parencite{roberts2025zerobench} \\
MMSI-Bench~\parencite{yang2025mmsibench} & MMStar~\parencite{chen2024we} & ScreenSpot-v2~\parencite{wu2024osatlas} \\
VisuLogic~\parencite{xu2025visulogic} & MME-CC~\parencite{zhang2025mmecc} & MathVista~\parencite{lu2024mathvista} \\
OSWorld-G~\parencite{xie2025scaling} & CharXiv-Reasoning~\parencite{wang2024charxiv} & MathVerse~\parencite{zhang2024mathverse} \\
MMMU-Pro~\parencite{yue2025mmmupro} & MMMU~\parencite{yue2024mmmu} & VPCT~\parencite{brower2025vpct} \\
SimpleVQA~\parencite{cheng2025simplevqa} & TreeBench~\parencite{wang2025treevgr} & MathKangaroo~\parencite{cherian2024evaluating} \\
DA-2K~\parencite{yang2024depth} & ZeroBench-sub~\parencite{roberts2025zerobench} & CountBench~\parencite{paiss2023teaching} \\
DynaMath~\parencite{zou2024dynamath} & RealWorldQA~\parencite{xai2024realworldqa} & RefSpatial-Bench~\parencite{zhou2025roborefer} \\
\bottomrule
\end{tabular}
}
\vspace{0.2cm}
\caption{The 42 source benchmarks aggregated for failure-driven capability discovery.}
\label{tab:source_benchmarks}
\end{table}

\label{sec:license}
All 42 source benchmarks aggregated in our failure-attribution study are publicly available research artifacts, and we use them in accordance with their original licenses.
Below, the two ZeroBench splits and the two CharXiv splits are counted under their parent benchmarks.
The majority are released under permissive licenses: Apache-2.0 (MMMU, MMMU-Pro, BLINK, ScreenSpot-v2, OSWorld-G, SimpleVQA, TreeBench, DA-2K, DynaMath, RefSpatial-Bench, LogicVista, VisFactor, MathCanvas, ViVerBench, VisuLogic), MIT (ChartQAPro, PhyX, VLMBias, ScreenSpot-Pro, VLMs-are-Blind, BabyVision, ZeroBench, MathVerse, VPCT, MathKangaroo, OCRBench-v2), or BSD-3-Clause (HallusionBench).
Nine benchmarks use Creative Commons licenses that require attribution: CC-BY-4.0 (ERQA, MuirBench, MMSI-Bench, MME-CC, CountBench) or CC-BY-SA-4.0 (MathVista, MathVision, MMStar, EMMA, and CharXiv).
Two benchmarks carry restrictive terms: We-Math (CC-BY-NC-4.0) and RealWorldQA (CC-BY-ND-4.0).
PointBench ships without an explicit license; we use it strictly for non-commercial research.
Every sample in \ourname carries provenance metadata (its source benchmark and original index), and source-derived samples remain subject to the licenses of their origin benchmarks.
RealWorldQA is used only for failure analysis and does not contribute any samples to \ourname.
Our own annotations and newly authored questions are released under Apache-2.0.

\section{Pairwise Overlap Between Source Benchmarks}
\label{sec:jaccard}

\begin{figure}[!htbp]
    \centering
    \includegraphics[width=0.82\textwidth]{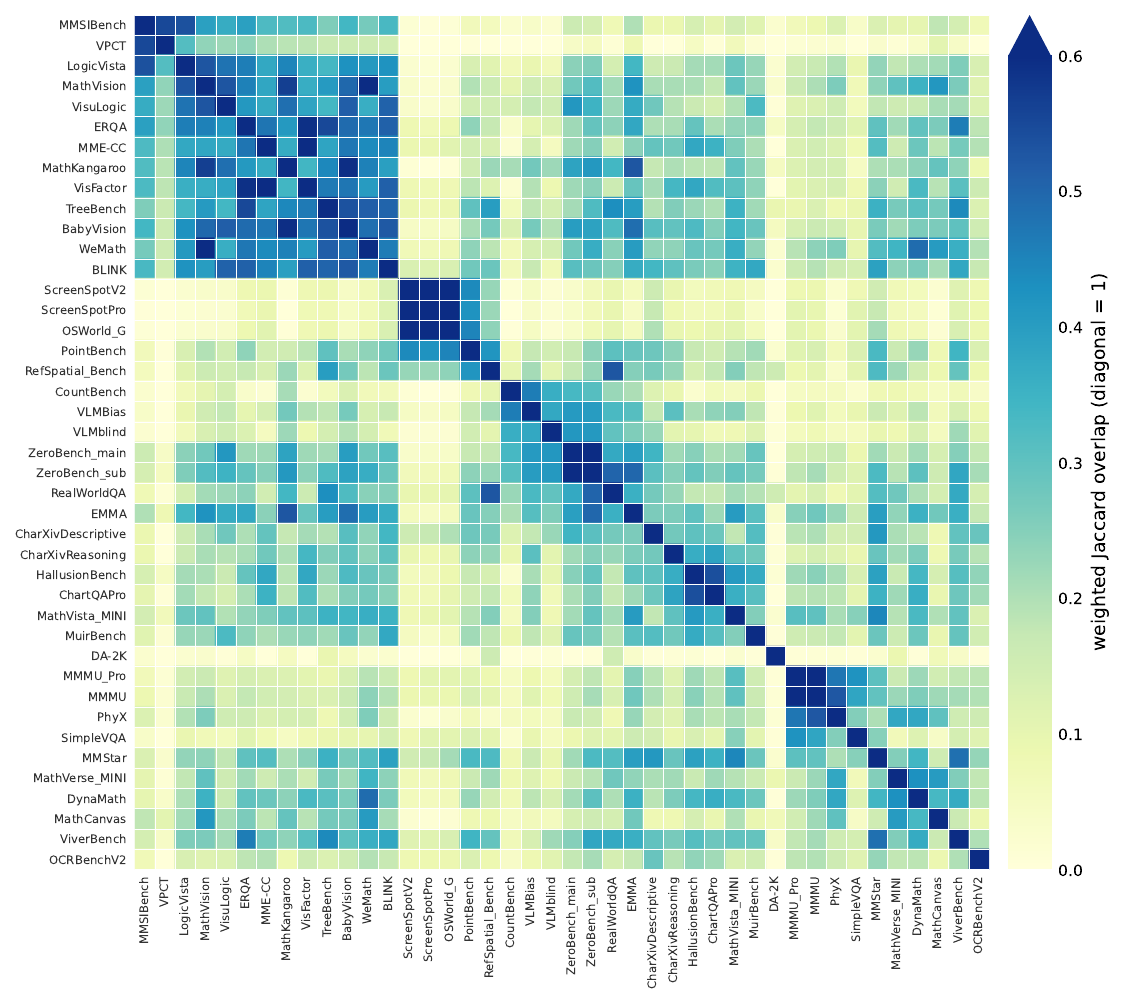}
    \caption{\textbf{Pairwise weighted Jaccard overlap between the error-type distributions of the 42 aggregated benchmarks.} The mean off-diagonal overlap is $0.20$; off-diagonal clusters are limited to same-family splits and same-task groups.}
    \label{fig:jaccard}
\end{figure}

Figure~\ref{fig:jaccard} reports the pairwise weighted Jaccard overlap between the error-type distributions of the 42 source benchmarks (benchmarks are ordered as in Figure~\ref{fig:coverage}).
The mean off-diagonal overlap is only $0.20$, and the few visible off-diagonal clusters correspond to near-duplicate splits of the same benchmark family (e.g., ScreenSpot-v2/ScreenSpot-Pro, ZeroBench-main/ZeroBench-sub, CharXiv-Reasoning/CharXiv-Descriptive) or to groups built around the same task (e.g., math word problems), while overlap across distinct families remains low.
This indicates that existing benchmarks provide fragmented and weakly-overlapping views of model failures, and that aggregating them recovers complementary slices of the failure space.

\section{Evaluation Prompt}
\label{sec:prompt}

\begin{tcolorbox}[
title=Judge Prompt,
colback=white,
colframe=black!70,
arc=5pt,
fonttitle=\bfseries]
Please act as a professional teacher and grade the student's answer. Below are the question, the student's answer, and the reference answer. Based on the question and the reference answer, analyze the student's answer and judge whether it correctly answers the question. I will provide several examples to help you understand how to judge.

========== [Output Format] ==========

Return your judgment in the following format. The first field [reason] gives the reason for your judgment; the second field [judge] gives your verdict as a single boolean, i.e., True or False. Make sure your output ends with True or False.

[reason]

<a brief justification of no more than 100 tokens>

[judge]

False

========== [Notice] ==========

\texttt{\{NOTICE\}}

========== [Examples] ==========

\texttt{\{EXAMPLES\}}

========== [Notice] ==========

\texttt{\{NOTICE\}}

Now you may begin grading.

========== [Question] ==========

\{problem\}

========== [Student Answer] ==========

\{assistant\_answer\}

========== [Reference Answer] ==========

\{reference\_answer\}

========== [Your Judgment] ==========
\end{tcolorbox}

\begin{tcolorbox}[
title=Judge Rules (\texttt{\{NOTICE\}}),
colback=white,
colframe=black!70,
arc=5pt,
fonttitle=\bfseries]
0. You only need to compare the consistency between the reference answer and the student's answer, focusing especially on the content after summarizing phrases such as "Final answer:". The reference answer is absolutely correct; judge solely by whether the student's final result matches it, even if the solution process is correct.

1. If the question contains multiple sub-questions, output the student's answer, the reference answer, and the consistency for each sub-question in [reason]; judge correct only when all sub-questions are consistent.

2. For multiple-answer questions, the student's answer must contain all correct answers without any extra ones. Pay special attention when the reference answer contains "or": decide whether all answers must be given based on the specific question.

3. If you believe the student's answer equals the reference answer after simplification, you must provide the complete simplification process in [reason] until the two are equal; otherwise it cannot be judged correct. You may not vaguely claim that "they can be made equal".

4. For numerical answers, integers or values with at most 4 significant figures must match exactly; values with more than 4 significant figures must match within 4 significant figures. In [reason], convert both to standard scientific notation, first compare the order of magnitude, then compare the first 4 significant figures. If the units differ, convert to a common unit first.

5. For English writing questions, if the student does not answer in English, judge it incorrect.

6. For physics, chemistry, and biology questions, if the reference answer contains a technical term, the student's answer must contain that exact term; synonyms are not accepted.

7. When the question asks to explain a term, redundant explanation is not penalized, but missing key points must be judged incorrect.

8. For multiple-choice questions, judge incorrect whenever the student's selected option differs from the reference answer, regardless of the content.
\end{tcolorbox}

\section{\ourname Showcases}
\label{sec:examples}

\begin{figure}[!htb]
\centering
\setlength{\tabcolsep}{2pt}

\newcommand{\exblock}[4]{%
\begin{minipage}[t]{0.48\textwidth}
\centering
\includegraphics[width=\linewidth,height=2.55cm,keepaspectratio]{#1}

\vspace{1mm}

{\scriptsize\raggedright
\centerline{\textbf{#2}}
\textbf{Question:} #3\\
\textbf{Answer:} #4\par
}
\end{minipage}
}

\exblock{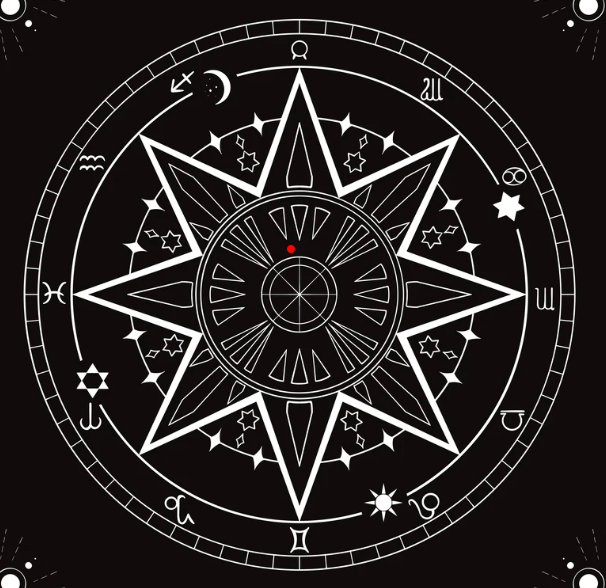}
{Visual Localization}
{At what o'clock position on the dial is the Gemini symbol in the image? Answer with just the number.}
{6}
\hfill
\exblock{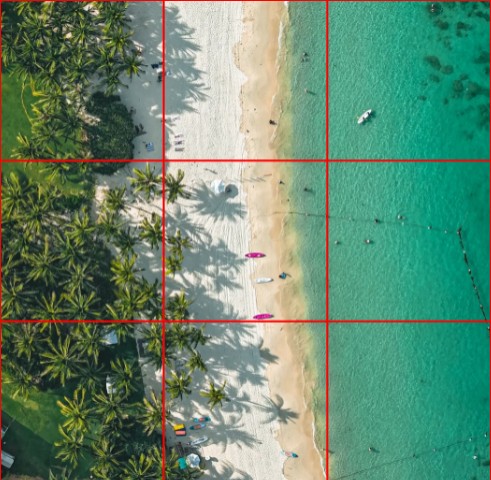}
{Visual Localization}
{The red lines in the image divide the picture into nine sections, which are numbered as regions 1--9 in order from left to right and then from top to bottom. Which of the following regions contains no trees at all?\\ A. Region 1\\ B. Region 2\\ C. Region 5\\ D. Region 8}
{B}

\vspace{3mm}

\exblock{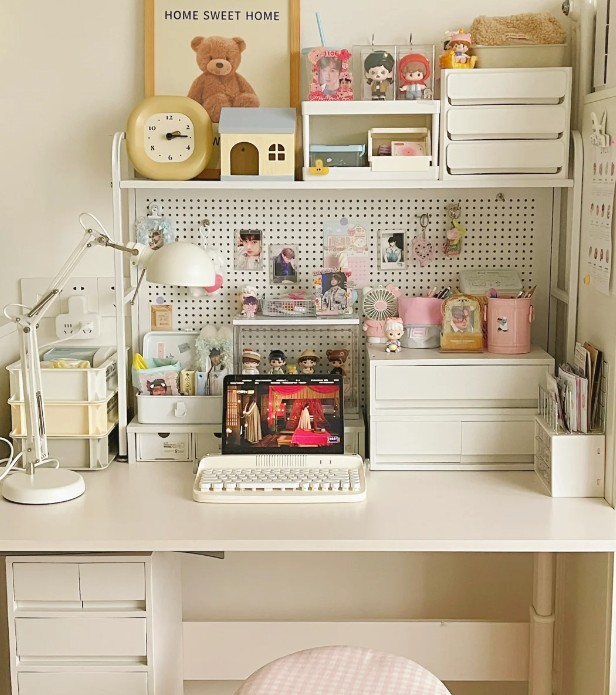}
{Visual Attribute}
{From the perspective shown in the image, compare the two pen holders with pink on the table. Which of the following conclusions is correct?\\ A. The left pen holder is pure pink with a cartoon character pattern on its surface; the right pen holder is gray-pink patchwork\\ B. The left pen holder is gray-pink patchwork; the right pen holder is pure pink with a cartoon character pattern on its surface\\ C. The left pen holder is pure pink; the right pen holder is gray-pink patchwork with a cartoon character pattern on its surface\\ D. The left pen holder is gray-pink patchwork with a cartoon character pattern on its surface; the right pen holder is pure pink}
{B}
\hfill
\exblock{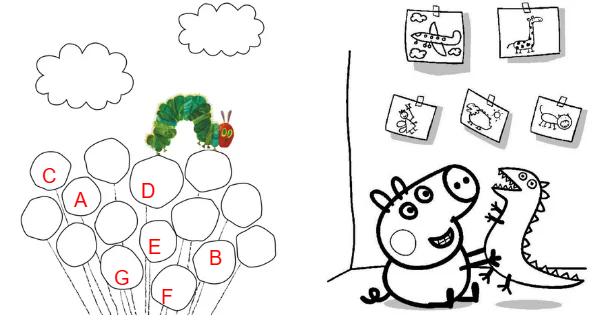}
{Visual Attribute}
{Observe the two cartoon characters in the lower right corner of the picture. Viewing from the perspective presented in the image, which statement correctly describes the composition of these two characters?\\ A. Both characters are composed entirely of curves\\ B. Both characters are composed of a combination of straight lines and curves\\ C. Both characters are composed entirely of straight lines\\ D. The character on the left is composed entirely of straight lines, while the character on the right is composed entirely of curves}
{B}

\vspace{3mm}

\exblock{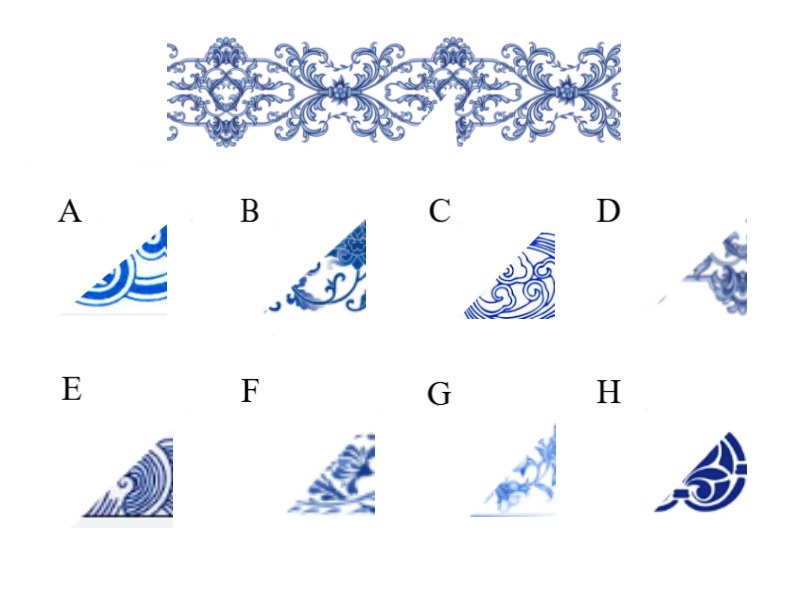}
{Fine-grained Recognition}
{Observe the outline shape of the notch in the main pattern above. From the eight options A, B, C, D, E, F, G, H, which is the only option whose edge contour can perfectly match the notch in the main body and seamlessly fill the missing area?}
{D}
\hfill
\exblock{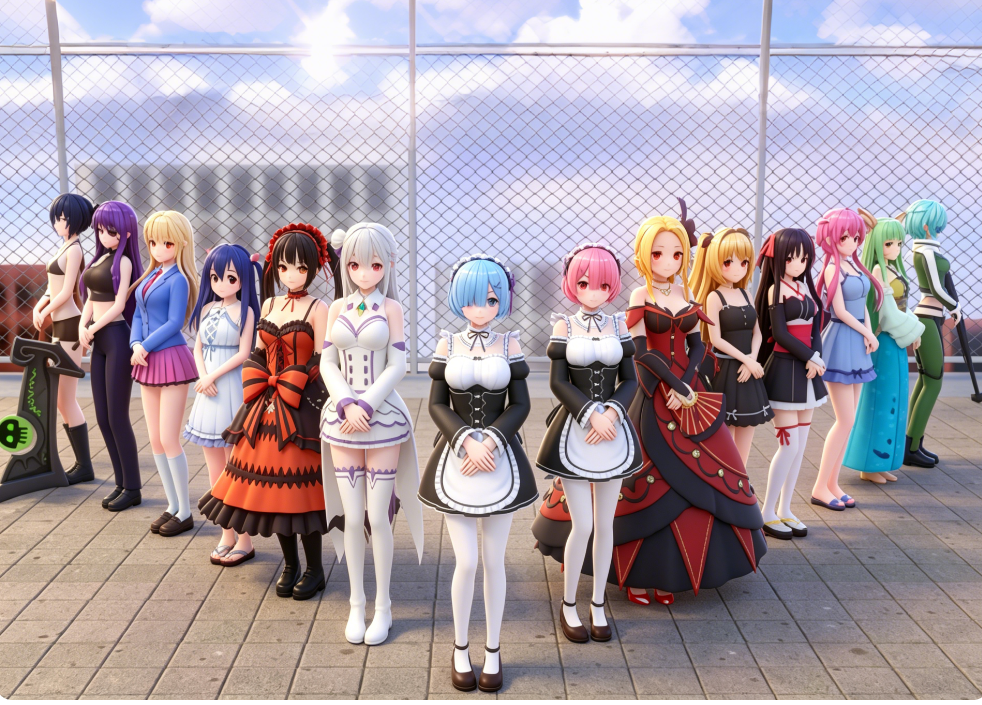}
{Fine-grained Recognition}
{Observe the image and compare the styling and outfits of the person in the center position (C position) with the person immediately to their right. Which of the following statements is correct:\\ A. The two have different hair colors, and their clothing color schemes and decorations are exactly the same\\ B. The two have different hair colors, and their clothing color schemes and decorations are clearly distinct\\ C. The two have the same hair color, and their clothing color schemes and decorations are exactly the same\\ D. The two have the same hair color, and their clothing color schemes and decorations are clearly distinct}
{A}

\caption{Representative examples from the ten atomic perceptual capabilities in \ourname (1 of 3).}
\label{fig:examples}
\end{figure}

\begin{figure}[!t]
\centering
\setlength{\tabcolsep}{2pt}

\newcommand{\exblock}[4]{%
\begin{minipage}[t]{0.48\textwidth}
\centering
\includegraphics[width=\linewidth,height=2.55cm,keepaspectratio]{#1}

\vspace{1mm}

{\scriptsize\raggedright
\centerline{\textbf{#2}}
\textbf{Question:} #3\\
\textbf{Answer:} #4\par
}
\end{minipage}
}

\exblock{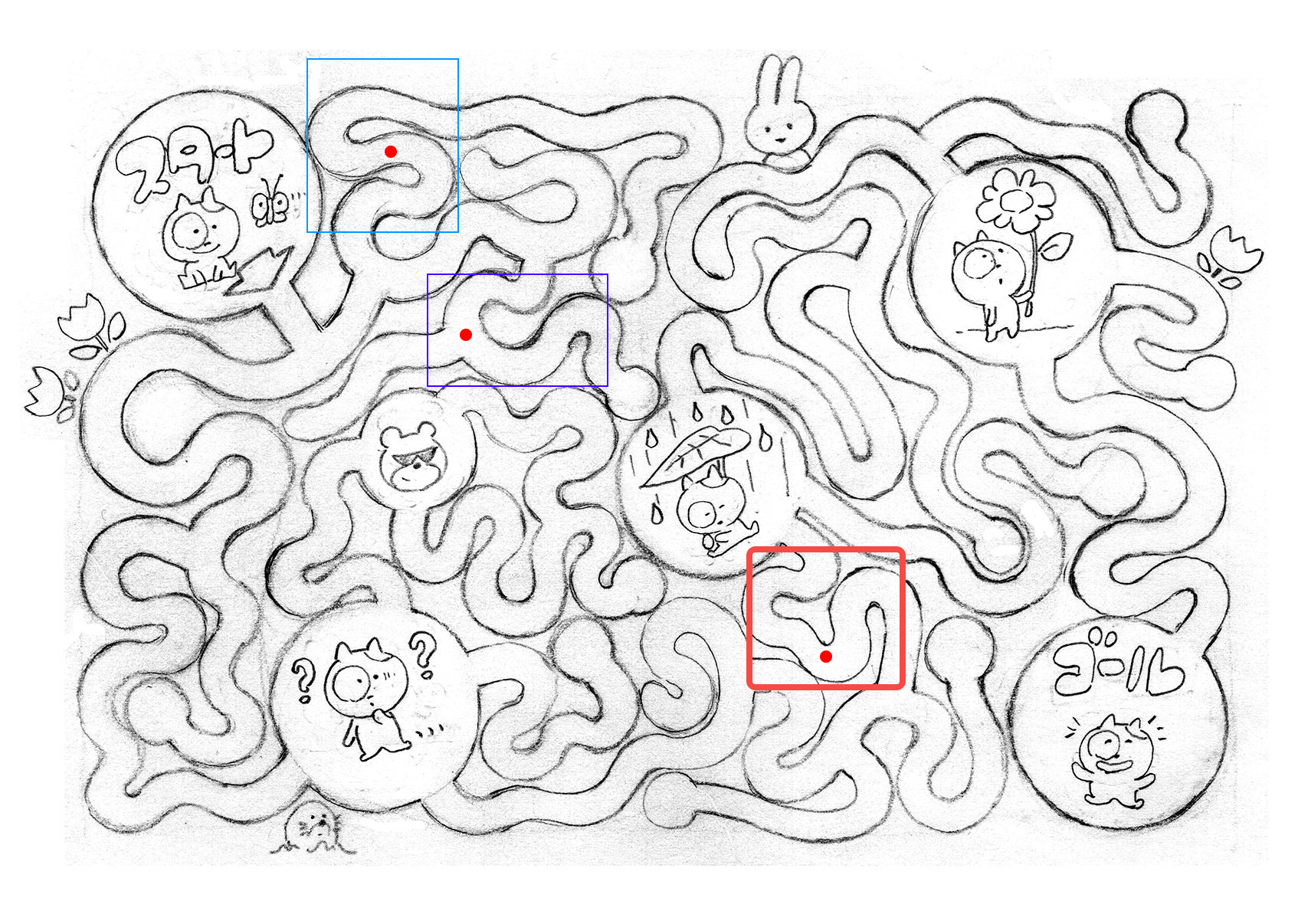}
{Visual Relation}
{Locate the solid dot inside the red box. As the dot travels along the existing route of the maze, which cat area does it reach first?\\ A. The puzzled cat at the lower left\\ B. The happy cat at the lower right\\ C. The cat holding an umbrella in the center\\ D. The cat admiring flowers at the upper right}
{C}
\hfill
\exblock{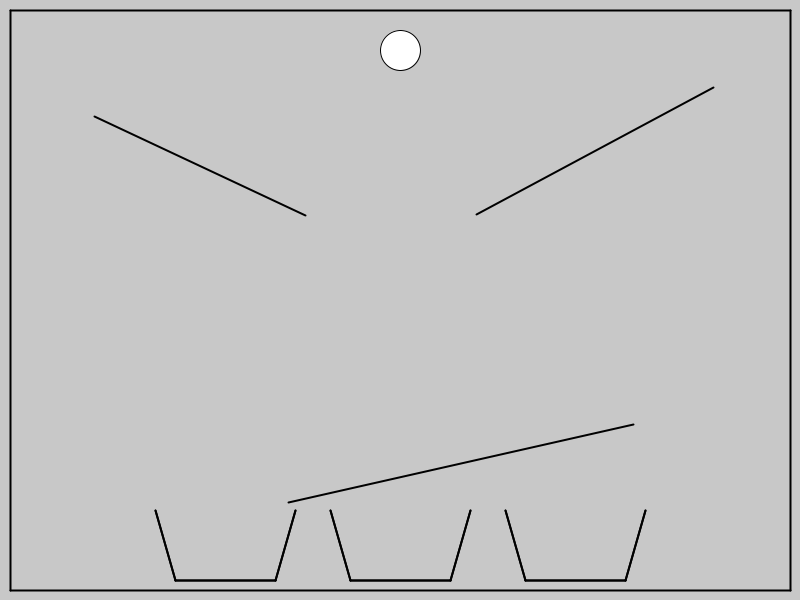}
{Visual Relation}
{In the figure, for the longest black diagonal line segment, in which direction does the line segment extending outward from its right endpoint point?\\ A. Upper right\\ B. Directly to the right\\ C. Lower right\\ D. Directly downward}
{A}

\vspace{3mm}

\exblock{}
{Depth \& 3D}
{Based on the current observer's perspective, with the direction closer to the camera defined as the front, between the person in red and the black electric fan in the center of the image, which one is further back?\\ A. The person in red\\ B. The black electric fan\\ C. Both are the same\\ D. Cannot be determined}
{A}
\hfill
\exblock{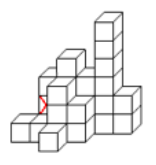}
{Depth \& 3D}
{If the small cubes stacked in the figure must be placed on the ground or on another small cube, how many small cubes are there in total in the figure (including the ones that cannot be seen)?}
{26 cubes}

\vspace{3mm}

\exblock{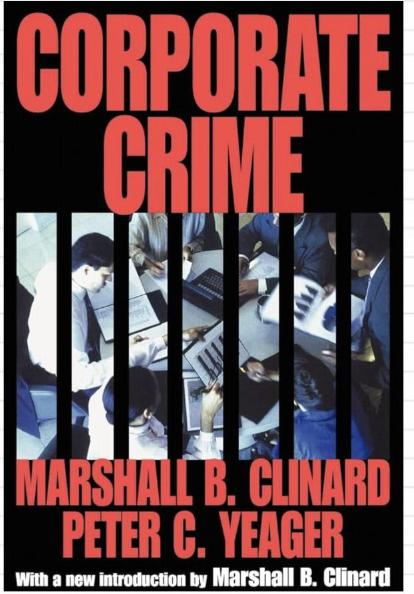}
{OCR}
{Observe the image. The white text appears in two different font sizes. For the white text in the largest font, read the text content from left to right, paying attention to uppercase and lowercase letters, and preserving all punctuation marks and spaces. Write down the answer.}
{Marshall B. Clinard}
\hfill
\exblock{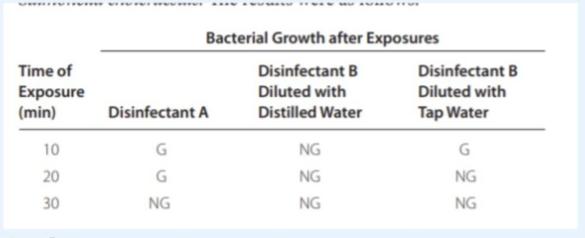}
{OCR}
{(Note: bounding boxes are given in [x1, y1, x2, y2] format.) Observe the image, establish normalized coordinates, read the text content in the coordinate region [0.697, 0.429, 0.875, 0.521] from left to right, distinguish case, and preserve spaces and punctuation, then write the answer.}
{Tap Water}

\caption{Representative examples from the ten atomic perceptual capabilities in \ourname (2 of 3).}
\label{fig:examples2}
\end{figure}

\begin{figure}[!t]
\centering
\setlength{\tabcolsep}{2pt}

\newcommand{\exblock}[4]{%
\begin{minipage}[t]{0.48\textwidth}
\centering
\includegraphics[width=\linewidth,height=2.55cm,keepaspectratio]{#1}

\vspace{1mm}

{\scriptsize\raggedright
\centerline{\textbf{#2}}
\textbf{Question:} #3\\
\textbf{Answer:} #4\par
}
\end{minipage}
}

\exblock{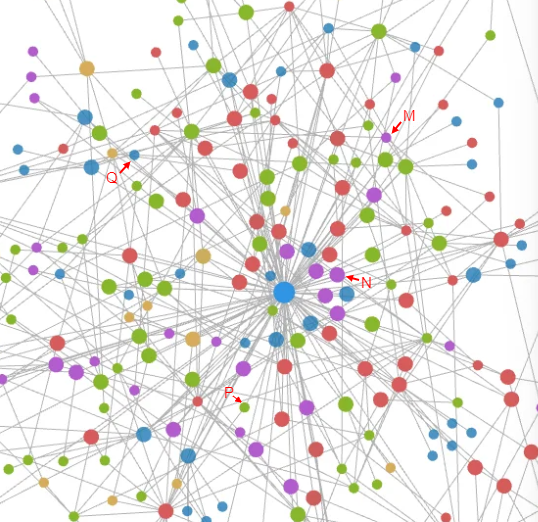}
{Visual Comparison}
{Among the circles corresponding to P, Q, M, and N, which is the largest circle? (Answer with the letter corresponding to the circle)}
{N}
\hfill
\exblock{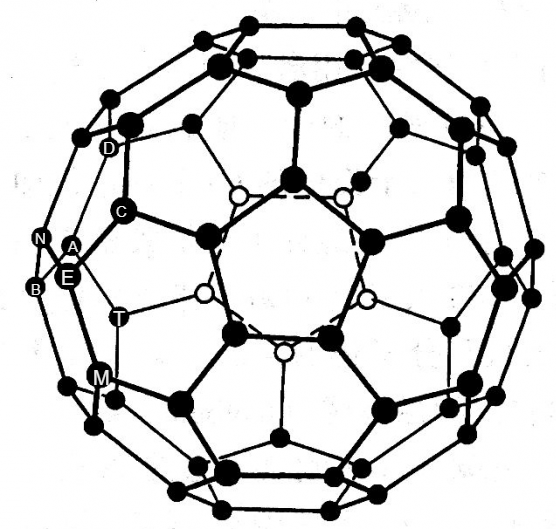}
{Visual Comparison}
{Let the line segment connecting data points A and D be denoted as segment AD, and let the line segment connecting data points E and M be denoted as segment EM. Which of the following is correct regarding the thickness of segments AD and EM?\\ A. Segment AD is thicker\\ B. Segment EM is thicker\\ C. Segments AD and EM are equally thick\\ D. Cannot be determined}
{B}

\vspace{3mm}

\exblock{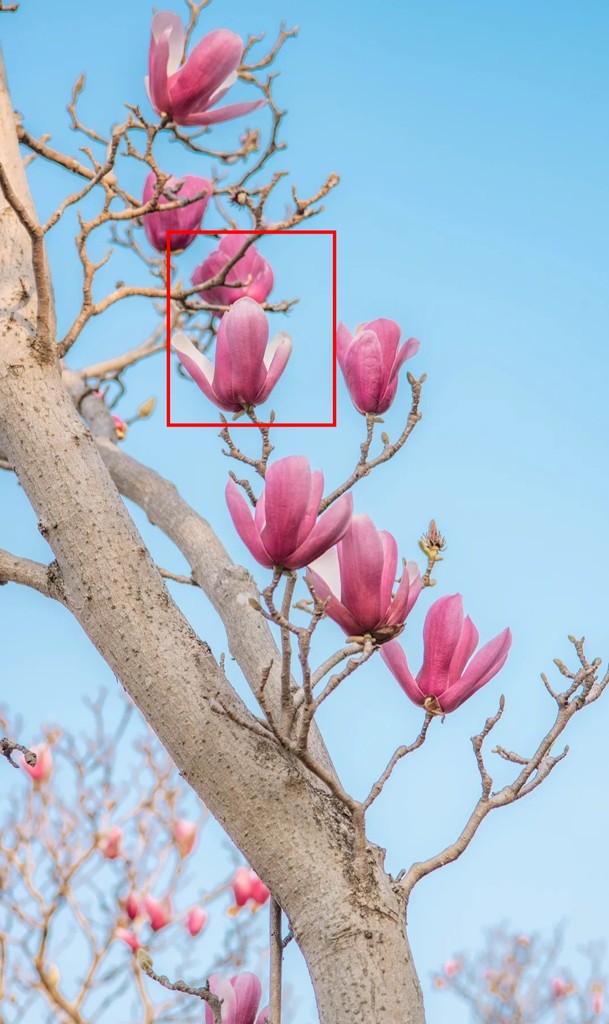}
{Visual Counting}
{Observe the red box in the picture. How many flowers have their main body inside the red box?}
{2}
\hfill
\exblock{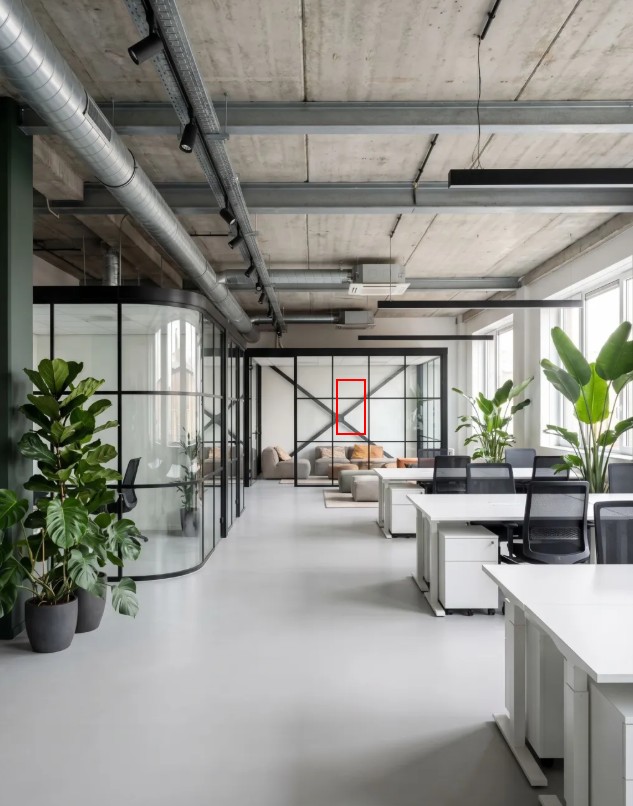}
{Visual Counting}
{How many potted green plants can be seen in the image in total? (excluding the green plants reflected in the glass)}
{5 pots}

\vspace{3mm}

\exblock{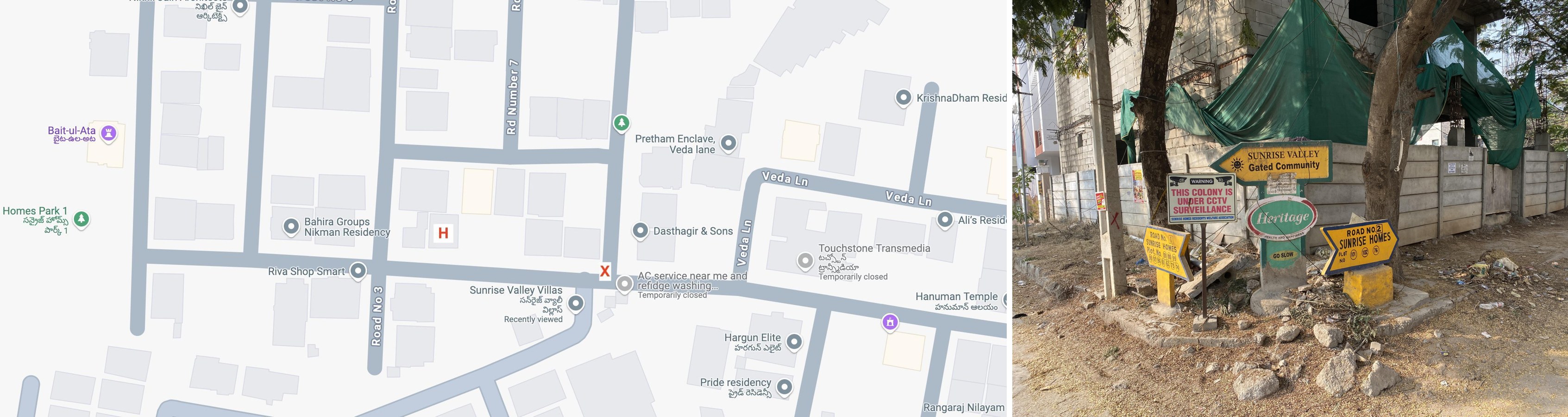}
{Context Integration}
{Please look at the map in the first image. I took another photo at the location marked with the red cross, which is the second image. Please determine which of the following buildings is on both sides of the road ROAD NO.2 SUNRISE HOMES.\\ A. Riva shop smart\\ B. Dasthagir \& Sons\\ C. Bait-ul-Ata\\ D. Sunrise Valley Villas}
{B}
\hfill
\exblock{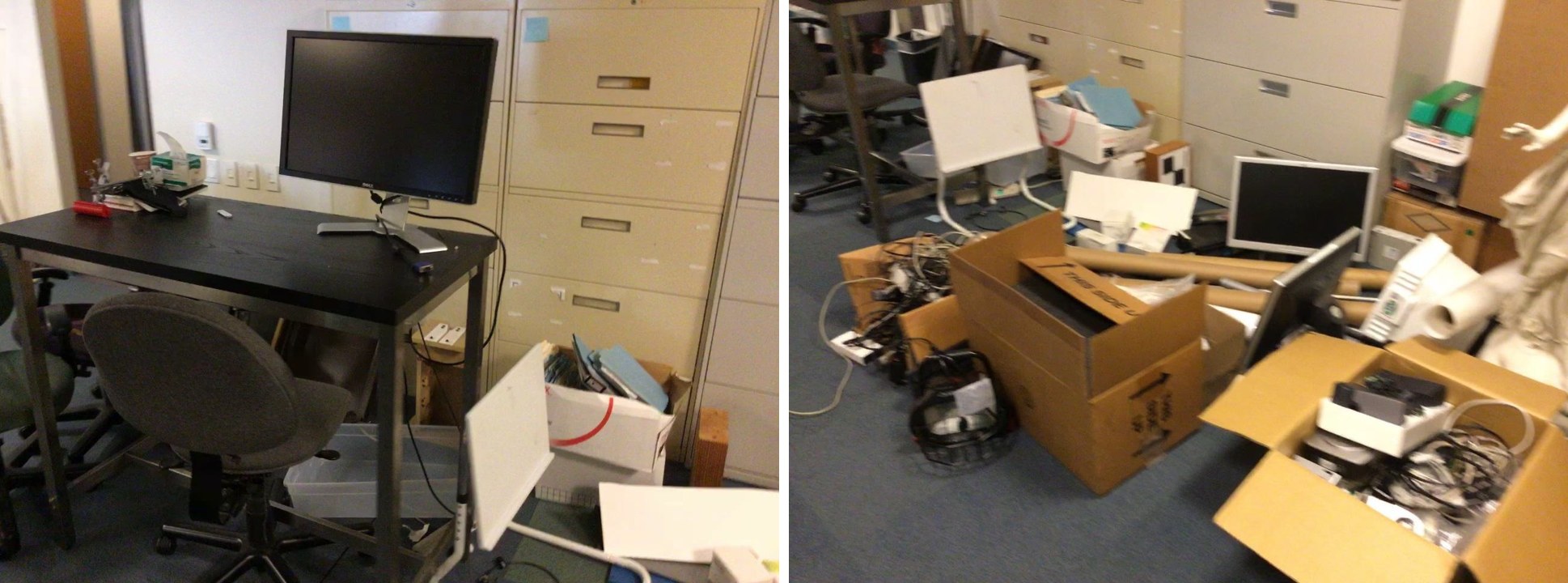}
{Context Integration}
{Compared with figure 1, what has been added to the scene in figure 2?\\ A. black office desk\\ B. semi-transparent storage basket\\ C. light yellow filing cabinet\\ D. cardboard box marked with an arrow}
{D}

\vspace{3mm}

\exblock{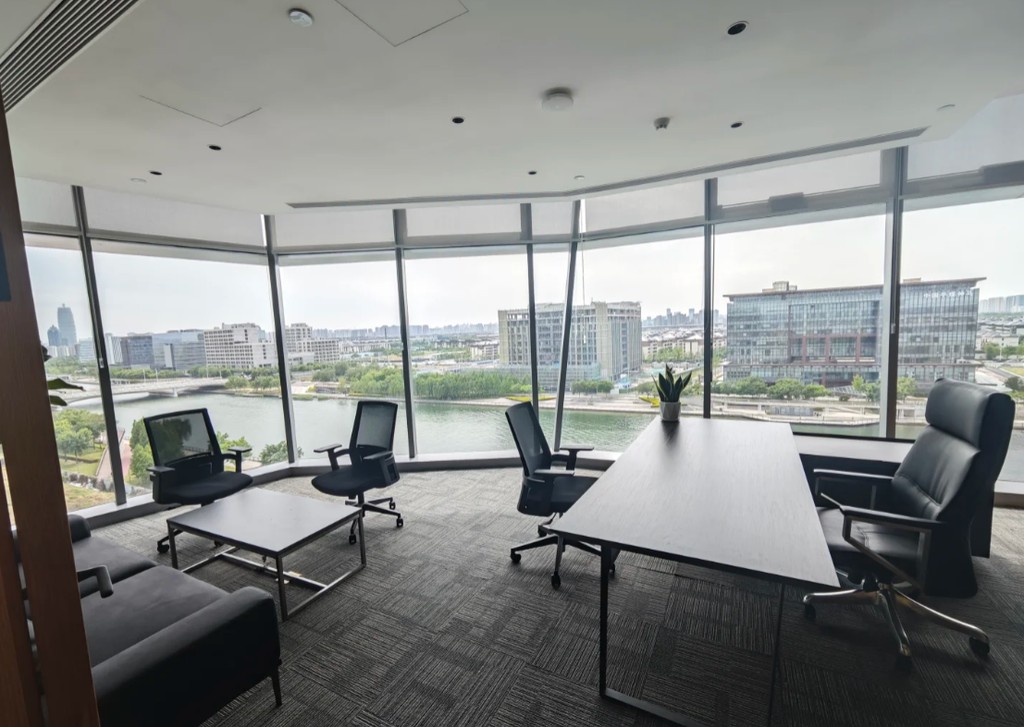}
{Hallucination}
{How many white tables are there in the office in the picture?\\ A. 0\\ B. 1\\ C. 2\\ D. 3}
{A}
\hfill
\exblock{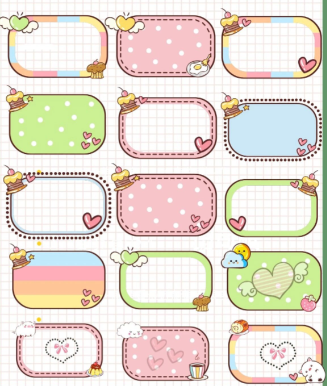}
{Hallucination}
{Among the 15 patterns in the image, how many patterns have a dog logo?}
{0}

\caption{Representative examples from the ten atomic perceptual capabilities in \ourname (3 of 3).}
\label{fig:examples3}
\end{figure}

Figures~\ref{fig:examples}--\ref{fig:examples3} present representative samples covering the ten atomic perceptual capabilities in \ourname, with the complete question text (including all answer options) shown for each sample.

\end{document}